\documentclass[lettersize,journal]{IEEEtran}
\usepackage[colorlinks,
          linkcolor=blue,
          anchorcolor=green,
          urlcolor=blue,
          citecolor=blue,
          ]{hyperref}
\usepackage{adjustbox}
\ifCLASSOPTIONcompsoc
\usepackage[caption=false, font=normalsize, labelfont=sf, textfont=sf]{subfig}
\else
\usepackage[caption=false, font=footnotesize]{subfig}
\fi
\usepackage{url}
\usepackage[normalem]{ulem}

\usepackage[flushleft]{threeparttable}
\usepackage{booktabs}
\usepackage{makecell}
\usepackage{multirow}
\usepackage{graphics}
\usepackage{graphicx}
\usepackage{algorithm}
\usepackage{algpseudocode}
\usepackage{amsthm,amsmath,amssymb}

\usepackage{mathrsfs}
\usepackage{cleveref}
\usepackage{cite}

\usepackage{subfig}
\def\BibTeX{{\rm B\kern-.05em{\sc i\kern-.025em b}\kern-.08em
    T\kern-.1667em\lower.7ex\hbox{E}\kern-.125emX}}
\usepackage{balance}

\newcounter{subeq}

\begin{document}
\title{Adversarial Training for Robust Coverage Network under Worst-case Facility Losses}

\author{Chang-Hao Miao, Yun-Tian Zhang, Tong-Yu Wu,\\ Fang Deng, \emph{Fellow}, \emph{IEEE} and Chen Chen, \emph{Member}, \emph{IEEE}\thanks{This work was supported by the National Key Research and Development Program of China No.2022ZD0119703; in part by the National Natural Science Foundations of China (NSFC) under Grant 62273044 and 62022015; in part by the National Natural Science Foundation of China National Science Fund for Distinguished Young Scholars under Grant 62025301; in part by the National Natural Science Foundation of China Basic Science Center Program under Grant 62088101. \emph{(Corresponding authors: Chen Chen)}}
\thanks{Chang-Hao Miao, Yun-Tian Zhang, Tong-Yu Wu, Fang Deng, and Chen Chen are with the State Key Laboratory of Autonomous Intelligent Unmanned Systems, Beijing Institute of Technology, Beijing 100081, China (e-mail: xiaofan@bit.edu.cn). Yun-Tian Zhang is also with the Department of Computer, Control and Management Engineering ``Antonio Ruberti", Sapienza University of Rome, Rome, 00185, Italy. Fang Deng and Chen Chen are also with the School of AI, Beijing Institute of Technology, Beijing 100081, China.}}

\markboth{\parbox[t]{\textwidth}{This work has been submitted to the IEEE for possible publication.\newline Copyright may be transferred without notice, after which this version may no longer be accessible.}}{How to Use the IEEEtran \LaTeX \ Templates}

\maketitle

\begin{abstract}
The Maximal Covering Location-Interdiction Problem (MCLIP) is a classic bi-level optimization problem, which is fundamental to resilient infrastructure planning yet remains computationally intractable. Specifically, the upper level determines facility locations to maximize coverage, while the lower level executes worst-case interdiction to minimize the coverage. The strong coupling between the upper and lower levels, combined with their respective high combinatorial complexity, renders traditional methods ineffective. To bridge this gap, we propose a Dual-Agent Deep Reinforcement Learning (DADRL) framework based on adversarial learning, comprising a location agent corresponding to the upper level and an interdiction agent corresponding to the lower level. Our contributions are threefold: (1) The location agent is trained simultaneously against an evolving interdiction agent, making it effectively capture the dynamic competitive interplay between the upper and lower levels; (2) To fully exploit the learned capabilities of the interdiction agent, we propose a Surrogate-based Ensemble Inference Strategy that utilizes the trained interdiction agent as a high-fidelity surrogate to guide the decisions of location agent; (3) Extensive experiments on synthetic and real-world datasets demonstrate that our approach achieves superior computational efficiency while maintaining highly competitive solution quality compared to other baselines. Furthermore, our DADRL framework is model-agnostic to network structures, while its underlying adversarial learning paradigm demonstrates strong potential for solving other bi-level optimization problems.
\end{abstract}


\section{Introduction}
\IEEEPARstart{C}{ritical} infrastructure networks, ranging from emergency medical services and telecommunication grids to supply chains, serve as the backbone of modern society. Traditionally, the design of such networks has relied on models like the Maximum Covering Location Problem (MCLP) \cite{daskin1983maximum,church1974maximal,schilling1993review}, which emphasizes placing facilities to maximize overall efficiency and minimize overlap. However, one can assume that infrastructure will work as intended for most of the time, but sometimes it will not because something has been disrupted \cite{church2018disruption}. Therefore, networks optimized solely for efficiency are inherently vulnerable to service disruptions, and the disruption of key facilities can lead to catastrophic reductions in service quality. For example, the Wall Street Journal \cite{smith2014us} noted that taking out just nine strategically selected substations across the United States could bring darkness to most of the country. Meanwhile, the real-world events \cite{berman2007facility}, such as the devastation of oil infrastructure by Hurricane Katrina or the disruption of mail sorting facilities during anthrax attacks, highlight the urgent need for resilience. Consequently, the focus of facility location research has shifted from static efficiency to designing robust networks against the risks associated with potential facility failures.


To ensure the reliable design for various facility location problems, researchers have extensively investigated network optimization under uncertainty, a progression systematically detailed in comprehensive reviews by Snyder \textit{et al.} \cite{snyder2016or} and Smith \textit{et al.} \cite{smith2020survey}. Researchers have attempted to enhance network reliability by modeling stochastic facility failures under various probabilistic assumptions \cite{drezner1987heuristic,lim2010facility,li2022general} or employing backup coverage models to ensure service redundancy \cite{hogan1986concepts,karatas2019analysis,tao2022location}. However, these studies are often insufficient for hedging against long-term failures or intelligent adversaries who exploit system vulnerabilities.

To address this challenge, O'Hanley \textit{et al.} \cite{o2011designing} introduced the Maximal Covering Location-Interdiction Problem (MCLIP), which explores strategies for coping with extreme conditions given the worst-case instances of long-term facility loss or destruction. Specifically, MCLIP integrates a deterministic interdiction framework into a minimax regret criterion, explicitly formulated as a Bi-level Optimization Problem (BOP). While the upper-level problem seeks to maximize the combination of initial and post-interdiction coverage, the lower-level interdiction framework is modeled as an $r$-interdiction covering problem (RIC) \cite{church2004identifying}, which identifies critical nodes in the facility network to cause maximum disruption.

Despite its practical significance, solving MCLIP poses severe computational challenges due to its intrinsic bi-level structure, rendering commercial solvers (e.g., CPLEX \cite{cplex}, Gurobi \cite{gurobi}) incapable of solving it directly. Although MCLIP can be formulated as a Mixed-Integer Program (MIP), this formulation relies on explicitly enumerating the entire set of feasible interdiction patterns, which may include a combinatorial number of constraints corresponding to each pattern. Regarding heuristic approaches for BOPs, most existing literature simplifies the lower-level problem into a greedy or approximate procedure to facilitate upper-level optimization. However, such simplifications often neglect the intrinsic strong coupling and dynamic competitive interplay between the upper and lower levels \cite{camacho2024metaheuristics}.


To bridge this gap, we propose a Dual-Agent Deep Reinforcement Learning (DADRL) framework based on adversarial training, comprising agents that correspond to the upper-level location and lower-level interdiction decisions. This work is the first to leverage Deep Reinforcement Learning (DRL) for such complex bi-level problems, which is also model-agnostic to network structures. Through simultaneous adversarial training, the location agent can capture the dynamic competitive interplay against evolving adversaries. Additionally, we propose a Surrogate-based Ensemble Inference Strategy that utilizes the trained interdiction agent as a high-fidelity surrogate to guide location decisions. Extensive experiments on synthetic and real-world datasets demonstrate that our approach offers superior efficiency and competitive optimality against baselines.

The rest of the paper is organized as follows. We briefly review the related works and illustrate our motivation in Section \ref{RW}. Section \ref{PS} details the mathematical formulation of MCLIP. Our methodology is explained in Section \ref{Methodology}. Section \ref{exp} presents simulations and experimental results. Section \ref{Conclusion} concludes this paper and discusses future work.

\section{Background and Motivation}
\label{RW}

\noindent In this section, we focus on the methodological developments in traditional approaches for bi-level optimization, as well as the application of DRL in combinatorial optimization. Based on this review, we identify the shortcomings of current methods and present the motivation for our work.

\subsection{Traditional Approaches for Bi-level Optimization}

BOPs model hierarchical decision processes involving two levels, described as mathematical programs
with an optimization problem embedded in their constraints \cite{bracken1973mathematical}. The intrinsic bi-level structure renders BOPs computationally intractable, with even the linear-linear case classified as Non-deterministic Polynomial Hard (NP-Hard) \cite{jeroslow1985polynomial}.

Real-world bi-level applications often involve complex integer or non-linear formulations, limiting the applicability of generic exact approaches. While reformulating the BOP into a single-level model using Karush-Kuhn-Tucker (KKT) conditions is common \cite{camacho2014solving,calvete2020AMF}, it is restricted to convex lower-level problems and hampered by Big-$M$ parameterization issues \cite{kleinert2020there}. Alternative exact techniques, such as vertex enumeration \cite{moore1990mixed}, parametric complementarity pivoting \cite{judice1988solution}, and branch-and-bound \cite{bard1990branch}, universally suffer from severe scalability limitations and fail to handle large-scale instances. For comprehensive reviews on exact methodologies, we refer readers to Saharidis \textit{et al.} \cite{saharidis2013exact} and Kleinert \textit{et al.} \cite{kleinert2021survey}.

To address the intractability of exact methods, metaheuristics have emerged as the dominant alternative for solving BOPs \cite{talbi2013taxonomy,camacho2024metaheuristics}. While some studies first reformulate the BOP into an equivalent single-level model before applying metaheuristics \cite{esfahani2022optimal,li2022hybrid,hayashi2023bilevel}, the vast majority of existing approaches adopt a nested architecture. In this paradigm, the upper level is typically guided by a metaheuristic, treating the lower level as a subordinate task that must be solved to optimality \cite{ziar2023efficient,peng2022research} or near-optimality \cite{lu2022bilevel,zhou2023bilevel} for every candidate solution generated. Specific variations, such as coevolutionary metaheuristics \cite{said2022discretization,chen2022integrated}, attempt to improve this process through the collaboration of upper and lower-level populations. Nevertheless, nested evaluation remains computationally prohibitive \cite{legillon2013cobra}, as it necessitates solving a nested optimization subproblem for thousands of iterations. To mitigate this burden, researchers have simplified the lower-level process using efficient techniques like greedy heuristics \cite{panin2014bilevel} or surrogate models \cite{mejia2020surrogate}. However, such simplifications reduce the lower-level problem to a static or approximated response, thereby failing to capture the strong coupling and dynamic competitive interplay intrinsic to the bi-level structure.


\subsection{DRL for Combinatorial Optimization Problems}

Beyond the routing problems, recent research has successfully adapted DRL to tackle the unique challenges of Facility Location Problems (FLPs), and initial efforts most focused on classical static variants. Chen \textit{et al.} \cite{chen2023attention} designed an attention model with multiple decoders to effectively balance spatial constraints in $p$-Center problem, while parallel works have targeted coverage maximization. For example, Wang \textit{et al.} \cite{wang2023deepmclp} introduced DeepMCLP to optimize urban spatial layouts, while Zhong \textit{et al.} \cite{zhong2024recovnet} proposed ReCovNet, a specialized architecture that explicitly embeds covering information into the state representation to solve the maximal coverage billboard location problem. 

Moving towards broader applicability and dynamic situations, Liang \textit{et al.} \cite{liang2024sponet} developed a unified framework capable of handling various spatial optimization problems. Furthermore, Miao \textit{et al.} \cite{miao2024deep} proposed ADNet, which integrates spatiotemporal features to solve the multi-period $pk$-median dynamic location problem, marking a shift from static snapshots to time-dependent decision-making. Meanwhile, the application of DRL has further extended to more complex Location-Routing Problems (LRPs), which requires considering location decisions and routing decisions simultaneously. For single-echelon systems, Miao \textit{et al.} \cite{miao2025end} presented a pioneering end-to-end learning approach for the Capacitated LRPs, effectively coordinating two interdependent decision stages. In hierarchical logistics networks, Huang et al. \cite{huang2025deep} adapted DRL specifically for the Two-Echelon Location-Routing Problem (2E-LRP), demonstrating the scalability of learning-based methods in handling multi-level constraints.

However, despite these significant strides, existing DRL frameworks predominantly focus on single-level optimization tasks, which lack the mechanisms to handle the bi-level adversarial constraints inherent in BOPs, including MCLIP.

\subsection{Motivation}

As discussed, exact methods fail to handle large-scale instances of MCLIP due to the inherent bi-level structure and the combinatorial explosion of interdiction patterns. While metaheuristics offer a trade-off between solution quality and computational efficiency, they frequently overlook the strong coupling and dynamic competitive interplay between the upper and lower levels. Furthermore, the nested architecture of existing metaheuristics inevitably compromises solution optimality, as the estimated lower-level reaction often deviates significantly from the true worst-case scenario. In contrast, DRL has demonstrated remarkable potential for solving combinatorial optimization problems with superior efficiency, yet its application has been predominantly confined to single-level optimization tasks.


Fortunately, MCLIP can be naturally described as a leader–follower or Stackelberg game \cite{simaan1973stackelberg}. Specifically, the leader locates $p$ facilities to maximize coverage, while the follower subsequently interdicts a subset $r$ of the facilities to minimize it. This interaction constitutes a competitive and adversarial process, in which the two parties pursue conflicting objectives. Inspired by the success of Generative Adversarial Networks (GANs) \cite{goodfellow2014generative} in modeling such minimax games, where a generator and a discriminator improve through mutual competition, we propose an adversarial DADRL framework for the MCLIP. In DADRL, we establish a location agent and an interdiction agent optimized against each other's evolving strategies. To further enhance the solution optimality, we also propose a Surrogate-based Ensemble Inference Strategy that utilizes the trained interdiction agent to guide the decisions of location agent.

To the best of our knowledge, this work represents the first application of DRL to BOPs, specifically targeting the MCLIP. Extensive experiments demonstrate that our DADRL framework achieves superior computational efficiency while maintaining highly competitive solution quality compared to other baselines.

\section{Problem Statement}
\label{PS}

\noindent In this section, we present the mathematical formulations of the MCLIP, based on O'Hanley \textit{et al.} \cite{o2011designing}. Specifically, we provide two equivalent formulation, including bi-level [MCLIP-1] and single-level [MCLIP-2] MIP formulation.


\subsection{Problem Description}

The MCLIP is naturally formulated as a bi-level optimization framework that seeks to locate $p$ facilities. The primary objective is to maximize the combination of coverage before and after the worst-case interdiction of $r$ facilities. The model operates under the assumption of homogeneous susceptibility, where any established facility is equally vulnerable to interdiction. Structurally, this problem integrates two distinct sub-problems: the upper level acts as an MCLP to establish the coverage network, while the lower level serves as an RIC to identify the specific subset of facilities whose removal causes the maximum loss in system coverage.


\subsection{Bi-Level MIP Formulation of MCLIP}

To formulate MCLIP as a bi-level MIP, we first define the sets, which are as follows:

\begin{itemize}
	\item $I$: Set of all customers.
	\item $J$: Set of all potential facility locations.
	\item $N_i$: Set of sites covering customer $i$.
\end{itemize}

\noindent We also provide the detailed illustration of all decision variables, as follows:

\begin{itemize}
	\item $x_j$: is equal to 1 if a facility is located at site $j$.
	\item $y_i$: is equal to 1 if customer $i$ is covered before interdiction.
	\item $u_j^\prime$: is equal to 1 if the facility at site $j$ is interdicted. 
	\item $y_i^\prime$: is equal to 1 if customer $i$ is covered after interdiction.
\end{itemize}

\noindent\textbf{[MCLIP-1]:}

\begin{small}
\begin{align}
\max_{x,y} \quad & z=\sum_{i\in I} y_i+z^\prime, \label{1}\\
\text{s.t.} \quad & \sum_{j \in J} x_j=p, \label{2}\\
& \sum_{j \in N_i} x_j\geq y_i \quad \forall i\in I, \label{3}\\
& \min_{u_j^\prime,y^\prime} \quad z^\prime=\sum_{i\in I}y_i^\prime, \label{4}\\
& \quad \text{s.t.} \quad \sum_{j\in J}u_j^\prime=r, \label{5}\\
& \phantom{\quad \text{s.t.} \quad} y_i^\prime+u_j^\prime\geq x_j \quad \forall i\in I, j\in N_i, \label{6}\\
& \phantom{\quad \text{s.t.} \quad} u_j^\prime\in\{0,1\} \quad \forall j\in J,\quad  y_i^\prime\geq 0\quad\forall i \in I,\label{7}\\
& x_j \in \{0,1\} \quad \forall j\in J,\quad y_i \leq1\quad\forall i\in I.\label{8}
\end{align}
\end{small}

\noindent The upper-level objective function (\ref{1}) aims to maximize the sum of covered customers before ($\sum_{i\in I} y_i$) and after $z^\prime$ interdiction. Constraint (\ref{2}) limits the number of open facilities to be equal to $p$. Constraint (\ref{3}) determines whether customer $i$ is covered by any open facilities before interdiction. Constraint (\ref{4}) defines the lower-level objective function, aiming to minimize the coverage after interdiction. Constraint (\ref{5}) requires the number of interdicted facilities to be equal to $r$. Constraint (\ref{6}) ensures that the customer $i$ is not covered only when its entire set of covering facilities is interdicted. Lastly, constraints (\ref{7}) and (\ref{8}) define the feasible region of the upper and lower level.

\subsection{Single-Level MIP Formulation of MCLIP}

While the bilevel formulation of MCLIP is conceptually straightforward, its intrinsic nested structure is intractable for standard commercial solvers designed for single-level MIPs. However, the problem can be reformulated into a single-level MIP by explicitly enumerating all feasible interdiction patterns. Through this transformation, the lower-level optimization is converted into a set of static constraints, but it inevitably introduces a combinatorial number of constraints. 

\begin{figure*}[!t]
	\centering
	\includegraphics[width=\textwidth]{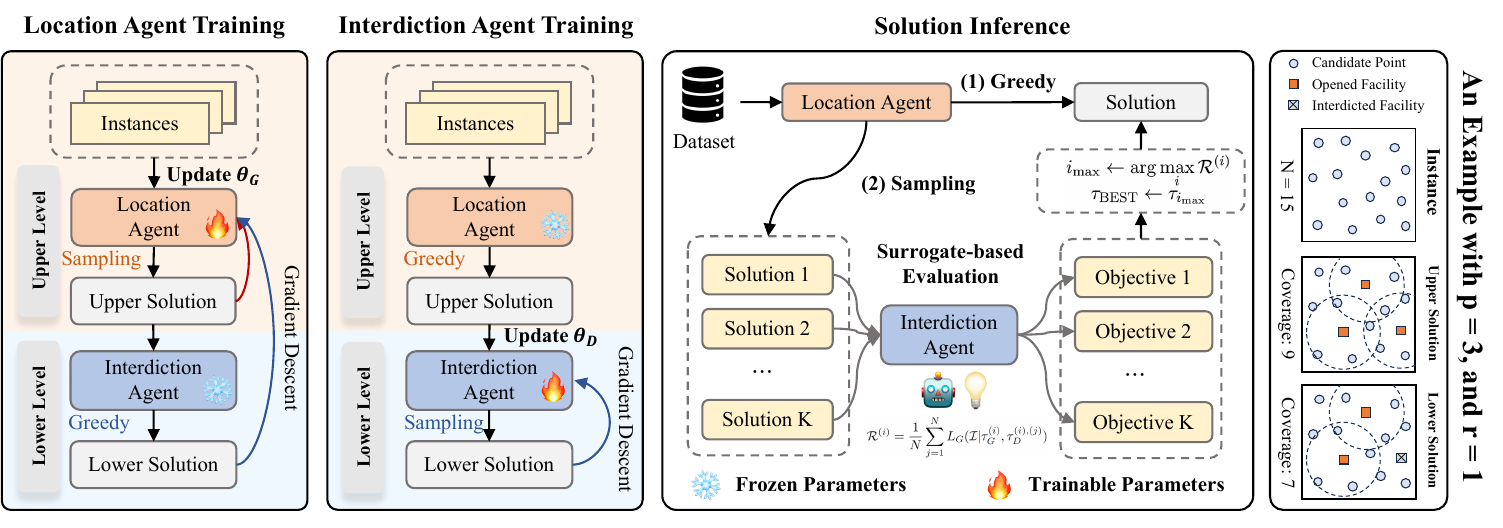}
	\caption{The overall framework of DADRL, which is model-agnostic to network structures. Our DADRL consists of three procedures: (1) Location Agent Training; (2) Interdiction Agent Training; and (3) Solution Inference. We also provide an example with $p=3$, and $r=1$, to better illustrate the problem.}
	\label{DADRL}
\end{figure*}

As to the single-level formulation of MCLIP, we have to explicitly define two new sets:
\begin{itemize}
	\item $K$: Set of numbered $p$ open facilities.
	\item $\Omega$: Set of all $\binom{p}{r}$ feasible interdiction patterns.
	\item $G_\omega$: Set of facility indices $k$ comprising interdiction $\omega$.
\end{itemize}

\noindent The decision variables can be defined as follows:

\begin{itemize}
	\item $x_{jk}$: is equal to 1 if facility $k$ is located at site $j$.
	\item $y_i$: is equal to 1 if customer $i$ is covered before interdiction.
	\item $y_{i\omega}^\prime$: is equal to 1 if customer $i$ is covered after interdiction.
	\item $z^\prime$: is the coverage under worst-case interdiction pattern.
\end{itemize}

\noindent\textbf{[MCLIP-2]:}

\begin{small}
\begin{align}
\max_{x,y} \quad & z=\sum_{i\in I} y_i+z^\prime, \tag{1}\\
\text{s.t.} \quad & \sum_{j \in J} x_{jk}=1\quad \forall k\in K, \label{9}\\
& \sum_{k \in K} x_{jk}\leq 1 \quad \forall j\in J, \label{10}\\
& \sum_{k \in K} \sum_{j \in N_i} x_{jk} \geq y_i \quad \forall i \in I, \label{11}\\
& \sum_{k\in K \setminus G_\omega} \sum_{j\in N_i}x_{jk}\geq y_{i\omega}^\prime\quad \forall \omega \in \Omega, i\in I,\label{12}\\
& \sum_{i\in I}y_{i\omega}^\prime\geq z^\prime\quad\forall \omega\in \Omega,\label{13}\\
& \sum_{j=1}^s x_{jk}\geq \sum_{j=1}^s x_{jk+1}\quad \forall k\in K,s\in J,\label{14}\\
& x_{jk}\in \{0,1\}\quad \forall j \in J, k \in K,\quad y_i \leq 1\quad \forall i \in I, \label{15}\\
& y_{i\omega}^\prime\leq1\quad \forall i \in I, \omega \in \Omega.\label{16}
\end{align}
\end{small}

\noindent The objective function (\ref{1}) remains to be the same as [MCLIP-1]. Constraint (\ref{9}) ensures that each facility $k$ chooses one site $j$ to locate at, thus making the total number of facilities to be equal to $p$. Constraint (\ref{10}) prevents the co-location situation for arbitrary two facilities. Constraint (\ref{11}) determines whether customer $i$ is covered by any open facilities before interdiction. Constraint (\ref{12}) ensures determines whether customer $i$ would be covered by any open, non-interdicted facilities. Constraint (\ref{13}) determines the minimum coverage over all possible interdiction patterns $\omega$. Constraint (\ref{14}) simply ensures that facility $k$ be located at a lower numbered site than facility $k + 1$, which can significantly reduce the solution space and is optional. Lastly, constraints (\ref{15}) and (\ref{16}) define the feasible region of decision variables.

\section{Methodology}
\label{Methodology}

\noindent In this section, we present the details of the DADRL framework. As illustrated in Fig. \ref{DADRL}, the framework is model-agnostic and features the simultaneous training of location and interdiction agents. The structure of this section is organized as follows: First, to facilitate the application of DRL algorithms, we reformulate the static upper-level location and lower-level interdiction problems into sequential Markov Decision Processes (MDPs), thereby constructing the respective agents. Next, we elaborate on the adversarial training pipeline, where these agents optimize against each other's evolving strategies. Finally, we propose a Surrogate-based Ensemble Inference Strategy that leverages the trained agents to significantly enhance solution quality during inference.

\subsection{Problem Reformulation as Markov Decision Processes}

\begin{figure}[!t]
	\centering
	\includegraphics[width=\columnwidth]{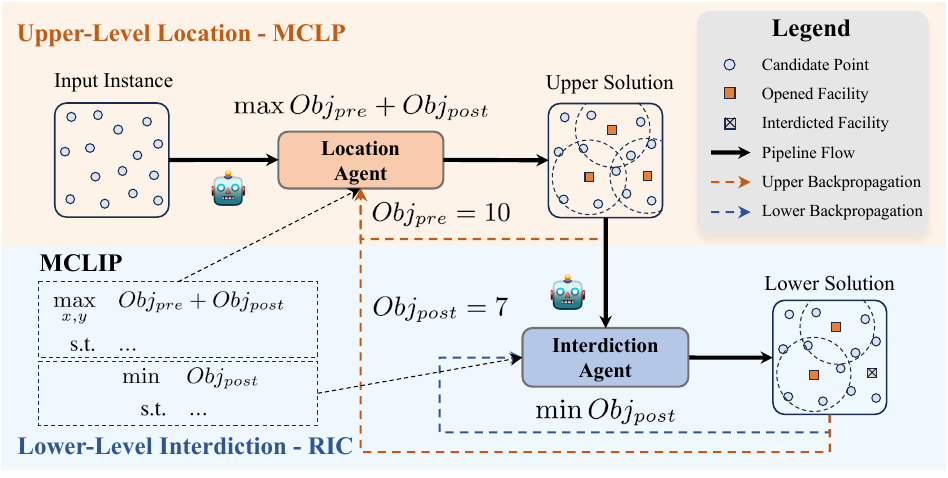}
	\caption{The interaction pipeline between location agent and interdiction agent.}
	\label{MDP}
\end{figure}

As mentioned before, the MCLIP comprises two nested optimization problems with conflicting objectives. Our DADRL framework addresses this by training an Upper-Level Location Agent and a Lower-Level Interdiction Agent simultaneously. By this way, we transform the strongly coupled bi-level problem into an interaction between two independent agents. The interaction pipeline is detailed in Fig. \ref{MDP}, where two agents have conflicting objectives and their respective actions and evaluations are inherently conditioned on the output of the counterpart.

Our framework considers the MCLIP as a dual-agent interaction, where the formulation of each single agent aligns perfectly with standard DRL works. To apply DRL to construct agents, we must reformulate the upper and lower-level problems as MDPs, each characterized by various state representations and action spaces.

\subsubsection{Upper-level Location}
For the Upper-level Location Agent, the problem is formulated as an MCLP, aiming to locate $p$ facilities to maximize the coverage while accounting for potential worst-case interdictions. Accordingly, we define the \textbf{state} as the spatial configuration of currently established facilities, and the \textbf{action} as the selection of a specific node from the candidate sites for location. Meanwhile, the \textbf{reward} can be directly defined as the sum of pre- and post-interdiction coverage $z$, according to the Eq. \ref{1}. Notably, the evaluation of post-interdiction coverage relies explicitly on the adversarial actions executed by the interdiction agent.

\subsubsection{Lower-level Interdiction}
Conversely, the Lower-level Interdiction Agent is formulated as an RIC. Its objective is to strategically disrupt a subset of $r$ facilities from the $p$ locations established by the location agent, specifically aiming to minimize the post-interdiction coverage. Therefore, the \textbf{state} is defined as the set of facilities currently selected for interdiction, while the \textbf{action} corresponds to the decision of interdicting a specific facility from the available set. To align with the minimization objective, we define the \textbf{reward} as the negative of post-interdiction coverage value $-z^\prime$, which can be computed based on the Eq. \ref{4}. Crucially, the decision space of interdiction  agent is inherently constrained by the location agent’s output, as it can only interdict facilities that have actually been deployed by location agent. 

\subsection{Dual-Agent Adversarial Training Mechanism}\label{subsec:training_mechanism}

To effectively handle the bi-level structure inherent in the MCLIP, we propose a dual-agent adversarial training mechanism for DADRL. Since the objective involves a Stackelberg game between the Location Agent (Leader) and the Interdiction Agent (Follower), standard independent reinforcement learning is insufficient. Instead, we adopt an iterative training scheme where the two agents are optimized alternately within each epoch to approximate the equilibrium. The overall procedure is outlined in Algorithm \ref{alg:training_mechanism}.

\subsubsection{Iterative Optimization Procedure}

The training process alternates between optimizing the Location Agent $\theta_L$ and the Interdiction Agent $\theta_I$. This decoupled approach ensures that each agent learns to react to an evolving counterpart:

\begin{itemize}
\item \textbf{Location Agent Training (Leader's Perspective):} When optimizing $\theta_L$, we aim to maximize the coverage performance assuming a rational adversary. For a given batch of instances $x$, the location decisions $\tau_L$ are sampled based on the location agent policy $\pi_{\theta_L}$ to encourage exploration. Crucially, the subsequent interdiction decisions $\tau_I$ is constructed greedily based on the interdiction agent policy $\pi_{\theta_I}$. This configuration simulates the worst-case scenario by enforcing a near-optimal response from the interdiction agent, while simultaneously preserving the exploration capability of the location agent to discover robust strategies.
\item \textbf{Interdiction Agent Training (Follower's Perspective):} Conversely, when optimizing $\theta_I$, we fix the location agent. The upper-level solutions $\tau_L$ are generated greedily to stabilize the environment for the follower. The interdiction decisions $\tau_I$ are then sampled from $\pi_{\theta_I}$ to explore diverse interdiction strategies against the fixed upper-level strategy.
\end{itemize}

As training progresses, both agents continuously explore to maximize their respective objectives, until the strategies converge to an equilibrium.


\begin{algorithm}[!t]
\caption{Adversarial Training Mechanism for DADRL}
\label{alg:training_mechanism}
\begin{algorithmic}[1]
\footnotesize
\State \textbf{Input:} Batch size $B$, Number of Epochs $N_{epoch}$;
\State \textbf{Output:} Trained Location Agent $\theta_L$ and Trained Interdiction Agent $\theta_I$;\\ 
\Comment{Initialization}\\
Initialize the parameters of both agents $\theta_L$, $\theta_I$;\\
Initialize the baseline policies $\theta_L^* \leftarrow \theta_L$, $\theta_I^* \leftarrow \theta_I$;\\
Generate validation dataset $\mathcal{X}_{val}$ for baseline policies;
\For{$epoch\leftarrow1:N_{epoch}$}
    \State Generate training dataset $\mathcal{X}_{train}$ for current epoch;
    \State \Comment{Location Agent Training}
    \For{each batch $x \in \mathcal{X}_{train}$}
    	\State Sample upper-level solutions $\tau_L \sim \pi_{\theta_L}(x)$;
    	\State Construct lower-level solutions greedily $\tau_I \sim \pi_{\theta_I}(x|\tau_L)$;
    	\State Construct baselines greedily $\tau_L^* \sim \pi_{\theta_L^*}(x)$, $\tau_I^* \sim \pi_{\theta_I^*}(x|\tau_L^*)$;
        \State Update Location Agent $\theta_L$ by REINFORCE:
        \[
        \resizebox{\linewidth}{!}{$
        \displaystyle
        \nabla \mathcal{L}_{\theta_L} = \frac{1}{B}\sum_{i=1}^B (R_{Loc}(x^{(i)}|\tau_L^{(i)},\tau_I^{(i)})-R_{Loc}(x^{(i)}|\tau_L^{*(i)},\tau_I^{*(i)}))\,\cdot\nabla_{\theta_L}\log \pi_{\theta_L}(\tau_L^{(i)})
        $}
        \]
    \EndFor
    \If {$\pi_{\theta_L}$ outperforms $\pi_{\theta_L^*}$ on validation dataset $\mathcal{X}_{val}$}
    	\State $\theta_L^* \leftarrow \theta_L$
    \EndIf
    \State \Comment{Interdiction Agent Training}
    \For{each batch $x \in \mathcal{X}_{train}$}
    	\State Construct upper-level solutions greedily $\tau_L \sim \pi_{\theta_L}(x)$;
    	\State Sample lower-level solutions $\tau_I \sim \pi_{\theta_I}(x|\tau_L)$;
    	\State Construct baselines greedily $\tau_L^* \sim \pi_{\theta_L^*}(x)$, $\tau_I^* \sim \pi_{\theta_I^*}(x|\tau_L^*)$;
        \State Update Interdiction Agent $\theta_{I}$ by REINFORCE:
        \[
        \resizebox{\linewidth}{!}{$
        \displaystyle
        \nabla \mathcal{L}_{\theta_I} = \frac{1}{B}\sum_{i=1}^B (R_{Int}(x^{(i)}|\tau_L^{(i)},\tau_I^{(i)})-R_{Int}(x^{(i)}|\tau_L^{*(i)},\tau_I^{*(i)}))\,\cdot\nabla_{\theta_I}\log \pi_{\theta_I}(\tau_I^{(i)})
        $}
        \]
    \EndFor
    \If {$\pi_{\theta_I}$ outperforms $\pi_{\theta_I^*}$ on validation dataset $\mathcal{X}_{val}$}
    	\State $\theta_I^* \leftarrow \theta_I$
    \EndIf
\EndFor
\end{algorithmic}
\end{algorithm}

\subsubsection{Gradient Estimation with Greedy Baselines}

To optimize the parameters of both agents, we employ the REINFORCE algorithm. A critical challenge in policy gradient methods is the high variance of the gradient estimate. To mitigate this, we introduce a self-critical baseline mechanism.

For the Location Agent, the gradient is approximated as:

\begin{equation}
\resizebox{.8\linewidth}{!}{$
\displaystyle
\nabla \mathcal{L}_{\theta_L} \approx \frac{1}{B}\sum_{i=1}^B \left( R_{Loc}(x^{(i)}|\tau_L^{(i)},\tau_I^{(i)}) - b_L(x^{(i)}) \right) \nabla_{\theta_L}\log \pi_{\theta_L}(\tau_L^{(i)})
$}
\end{equation}
where $R_{Loc}(\cdot)$ represents the reward $z$ for the location agent, which can be computed as Eq. \ref{1}.  The baseline term $b_L(x^{(i)})$ is obtained by executing both the current baseline policies $\pi_{\theta_L^*}$ and $\pi_{\theta_I^*}$ greedily, i.e., $b_L(x^{(i)}) = R_{Loc}(x^{(i)}|\tau_L^{*(i)},\tau_I^{*(i)})$. This baseline effectively centers the reward, reducing variance without introducing bias. A symmetric gradient estimation is applied for the Interdiction Agent using reward $R_{Int}(\cdot)$ and baseline policies, where $R_{Int}(\cdot)$ can be computed as $-z^\prime$ according to the Eq. \ref{4}.

\subsubsection{Baseline Update Strategy}

To ensure the baselines $\theta_L^*$ and $\theta_I^*$ provide a solid reference, they are not updated at every epoch. Instead, we maintain a separate validation dataset $\mathcal{X}_{val}$. At the end of each training phase within an epoch, we evaluate the current agent against their respective baseline on $\mathcal{X}_{val}$. The baseline parameters are updated ($\theta^* \leftarrow \theta$) only if the current policy outperforms the baseline policy. It should be mentioned, all solutions on validation dataset $\mathcal{X}_{val}$ are constructed greedily to exclude exploration noise during the evaluation.

\subsection{Surrogate-Assisted Inference Enhancement}

During the inference phase, solution generation can be typically categorized into two decoding strategies: (1) greedy decoding, which deterministically selects the action with the highest probability at each step; and (2) sampling decoding, which samples actions based on the learned probability distribution, and the best-performing solution is selected as the final solution. However, in the MCLIP, the evaluation of a solution relies on the optimal solution of the lower-level problem. Therefore, employing exact solvers (e.g., Gurobi) for this verification is computationally prohibitive, negating the speed advantage of DRL. Conversely, relying on greedy heuristics to approximate the lower-level response lacks precision, which leads to the incorrect identification of inferior solutions as the best, thereby degrading overall system performance.

\begin{algorithm}[H]
\caption{Surrogate-based Ensemble Inference Strategy}
\label{alg:inference}
\begin{algorithmic}[1]
\small
\State \textbf{Input:} Instance $\mathcal{I}$, Trained Location Agent $\theta_L$, Trained Interdiction Agent $\theta_I$, Sampling Size $K$, Ensemble Size $E$;
\State \textbf{Output:} Best-performing Solution of MCLIP $\tau_{BEST}$;\\ 
\Comment{Sampling by Location Agent}\\
Sample $K$ solutions $\tau^{(i)}_L \sim \pi_{\theta_L}(\mathcal{I})\quad \forall i \in \{1,...,K\}$;\\
\Comment{Evaluation by Interdiction Agent}\\
Sample $E$ times $\tau^{(i),(j)}_I \sim \pi_{\theta_I}(\mathcal{I}|\tau^{(i)}_L)\quad \forall i \in \{1,...,K\}, \forall j \in \{1,...,E\}$;\\
Calculate the Ensemble Reward:
\[
        \mathcal{R}^{(i)}=\frac{1}{E}\sum_{j=1}^ER_{Loc}(\mathcal{I}|\tau_L^{(i)},\tau_I^{(i),(j)})\quad \forall i \in \{1,...,K\}
\]\\
$i_{max}\leftarrow \arg\max_i \mathcal{R}^{(i)}$;\\
$\tau_{BEST}\leftarrow\tau_{i_{max}};$
\end{algorithmic}
\end{algorithm}

To address these limitations, we propose a Surrogate-based Ensemble Inference Strategy, which leverages the simultaneously trained Interdiction Agent as a high-fidelity, efficient surrogate evaluator. The details are outlined in Algorithm \ref{alg:inference}.

\begin{figure}[!t]
	\centering
	\includegraphics[width=.7\columnwidth]{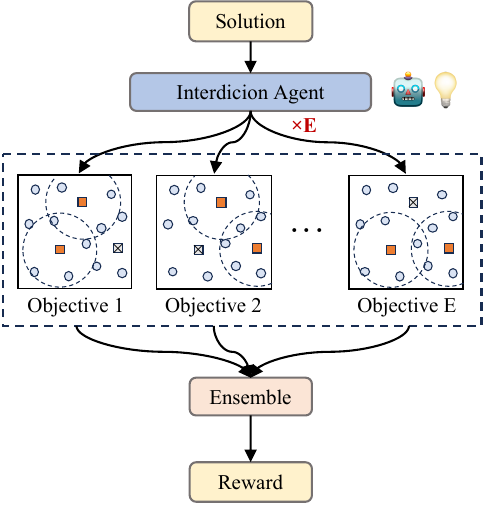}
	\caption{The illustration for surrogate evaluation via ensemble.}
	\label{Surrogate}
\end{figure}

\subsubsection{Candidates Generated via Sampling}

Instead of generating a single solution greedily, we leverage the stochastic nature of the trained Location Agent $\pi_{\theta_L}$ to generate a diverse set of candidate solutions. Specifically, we sample $K$ distinct location plans $\{\tau^{(1)}_L, \dots, \tau^{(K)}_L\}$ from the policy distribution $\pi_{\theta_L}$. This sampling process allows the agent to explore the solution space broadly, retrieving high-quality candidates that might be missed by a purely greedy search.

\subsubsection{Surrogate Evaluation via Ensemble}

Evaluating the performance of these $K$ candidates is computationally expensive if exact solvers are used to compute the optimal interdiction. Here, the trained Interdiction Agent $\pi_{\theta_I}$ serves as a high-fidelity surrogate for the lower-level problem. For each candidate solution $\tau^{(i)}_L$, we do not rely on a single interdiction response; instead, we sample $E$ potential interdiction patterns $\{\tau^{(i),(1)}_I, \dots, \tau^{(i),(E)}_I\}$ based on the Interdiction Agent's policy $\pi_{\theta_I}$. The evaluation fully exploit the trained interdiction agent and performs a reliable estimation of the worst-case loss.

\subsubsection{Selection of best-performing solution}

As shown in Fig. \ref{Surrogate}, we calculate the empirical expected reward for each candidate location plan based on the ensemble of simulated interdiction:

\begin{equation}
\mathcal{R}^{(i)} = \frac{1}{E}\sum_{j=1}^E R_{Loc}(\mathcal{I} \mid \tau_L^{(i)}, \tau_I^{(i),(j)})
\end{equation}

The candidate with the highest expected reward $\tau_{i_{max}}$ is selected as the final solution $\tau_{BEST}$. By integrating sampling-based exploration with surrogate-based evaluation, this strategy significantly boosts solution robustness and optimality with negligible computational overhead.

\section{Experimental Results}
\label{exp}
\noindent In this section, we conduct comprehensive experiments to evaluate the effectiveness of our proposed DADRL. We first detail the experimental settings and then compare our method against other representative baselines using synthetic datasets. Next, we evaluate the generalization performance across varying problem scales and on real-world datasets with distinct distributions. Additionally, we present ablation studies to analyze the contribution of each individual component.

\subsection{Experimental Settings}
\subsubsection{Datasets}

Our experiments are primarily conducted on two types of datasets: synthetic and real-world datasets. For the synthetic dataset, all candidate points are generated randomly within a unit square $[0\times1]^2$. We consider three problem scales with the number of nodes $N \in \{20, 50, 100\}$, and settings are detailed in Table \ref{scenario}. Since nodes are not uniformly distributed in real-world scenarios, we conduct comparative experiments on two real-world datasets:(1) The canonical facility location instances \footnote{\url{http://www.lac.inpe.br/~lorena/instancias.html}} from S\~ao Jos\'e dos Campos, Brazil (SJC), proposed by Pereira \textit{et al.} \cite{pereira2007column}; (2) A dataset we construct based on real-world data from Beijing, China (BJ), where each district serves as an instance containing up to 2,472 nodes.

 \begin{table}[h]
	\caption{Synthetic Datasets Settings.}
	\label{scenario}
    \centering
    \begin{tabular}{cccc}
    \toprule
        Settings & Small & Medium & Large \\ \midrule
        Graph size & 20 & 50 & 100 \\ 
        Number of facilities $p$ & 4 & 8 & 15 \\
        Number of interdiction $r$ & 1 & 3 & 5 \\
        Maximum service radius & 0.3 & 0.2 & 0.2 \\ \bottomrule
    \end{tabular}
\end{table}

\subsubsection{Training and Testing}

Given the model-agnostic nature of our DADRL framework, we select two representative DRL architectures as backbones to demonstrate its versatility: (1) AM \cite{kool2018attention}, a seminal work originally designed for generic routing problems; (2) ADNet \cite{miao2024deep}, An advanced extension of AM, which is explicitly tailored to capture the spatiotemporal features of facility location problems. All hyperparameters related to training and model architecture are kept the same with the related works, which are listed in Table \ref{hyperparameter}. Specifically, we set the initial learning rate to 1e-4 and is decayed by a factor of 0.1 every 200 epochs.

\begin{table}[h]
\setlength{\abovecaptionskip}{0.1cm}
	\caption{Hyperparameter Configurations.}
	\label{hyperparameter}
    \centering
    \begin{tabular}{cc}
    \toprule
        Hyperparameters & Value \\ \midrule
        Number of Epochs & 1,000 \\ 
        Number of Instances per Epoch & 512,000 \\
        Batch Size of Training & 512 \\ 
        Number of Instances per Evaluation & 1,280 \\
		Batch Size of Evaluation & 12,800 \\
		Hidden Dimension & 128 \\ 
		Embedding Dimension & 128\\
		Number of Attention Heads & 8\\
        Number of Encoder Layers & 3 \\ 
        Optimizer & Adam \\ 
        Learning rate & 1e-4 \\ \bottomrule
    \end{tabular}
\end{table}

\begin{figure}[t]
\centering
\setlength{\tabcolsep}{1pt}
\begin{tabular}{ccc}
\subfloat[Location agent of MCLIP20.\vspace{-2mm}]{\includegraphics[width=.49\linewidth]{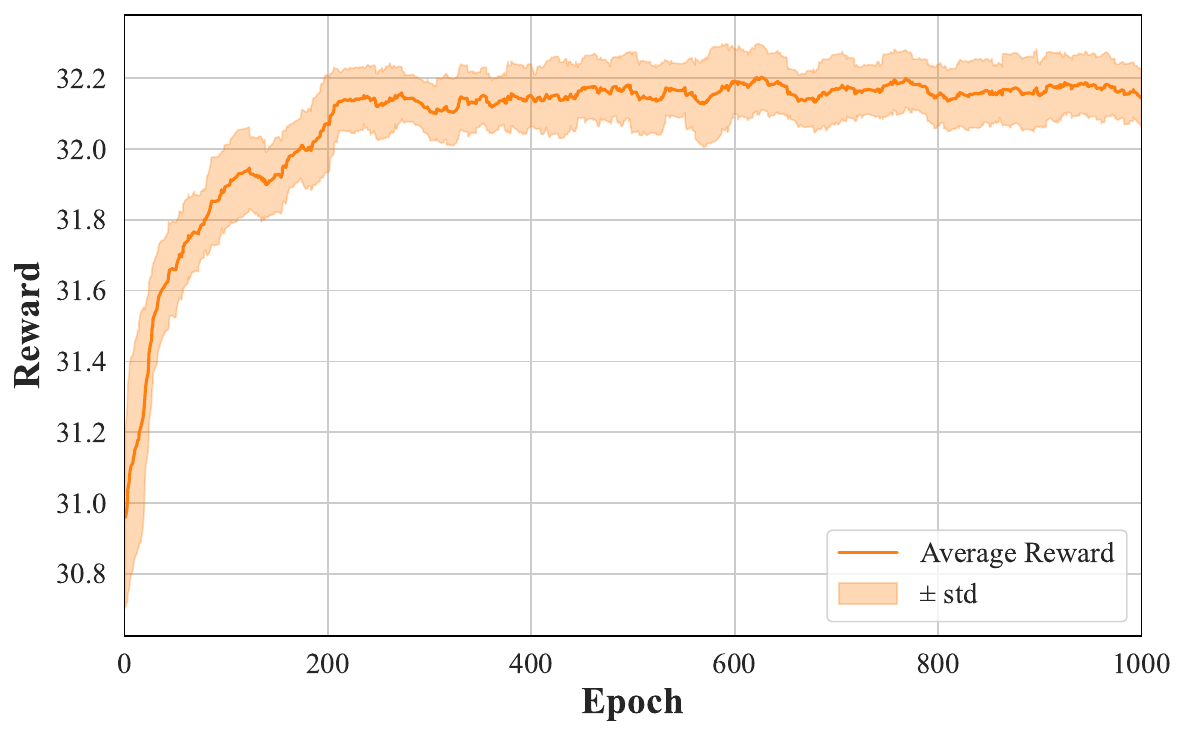}} &
\subfloat[Interdiction agent of MCLIP20.\vspace{-2mm}]{\includegraphics[width=.49\linewidth]{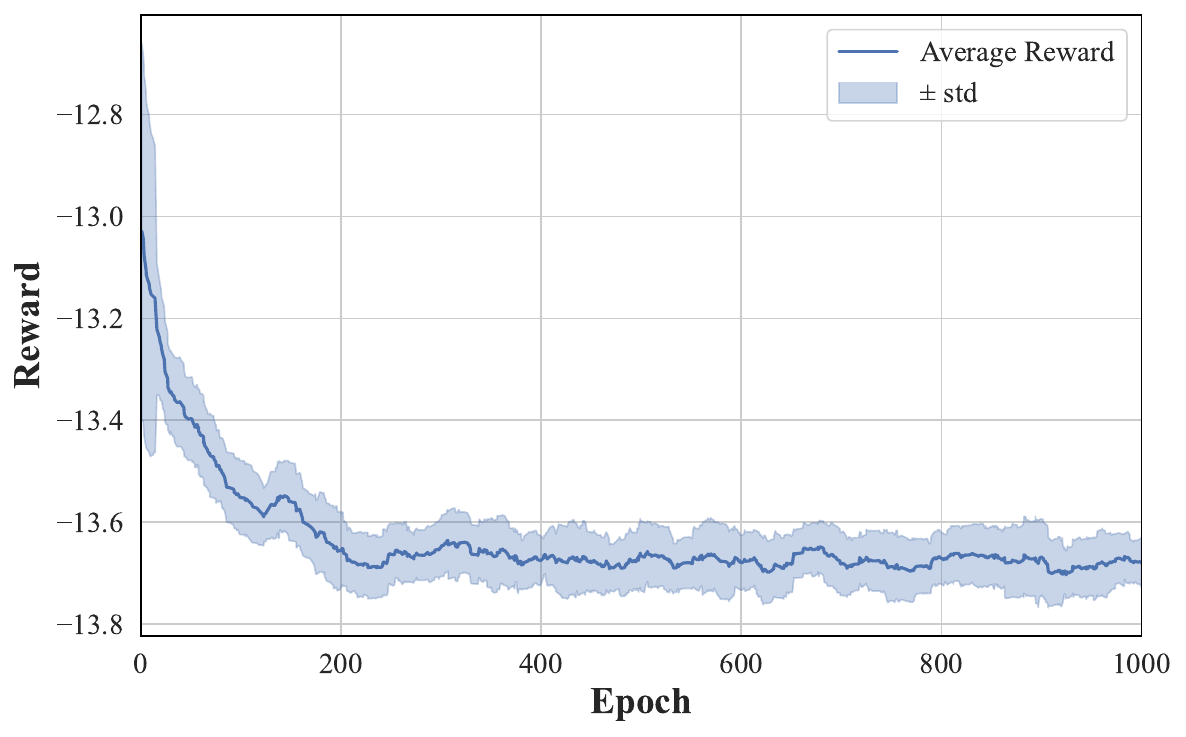}} &\\
\subfloat[Location agent of MCLIP50.\vspace{-2mm}]{\includegraphics[width=.49\linewidth]{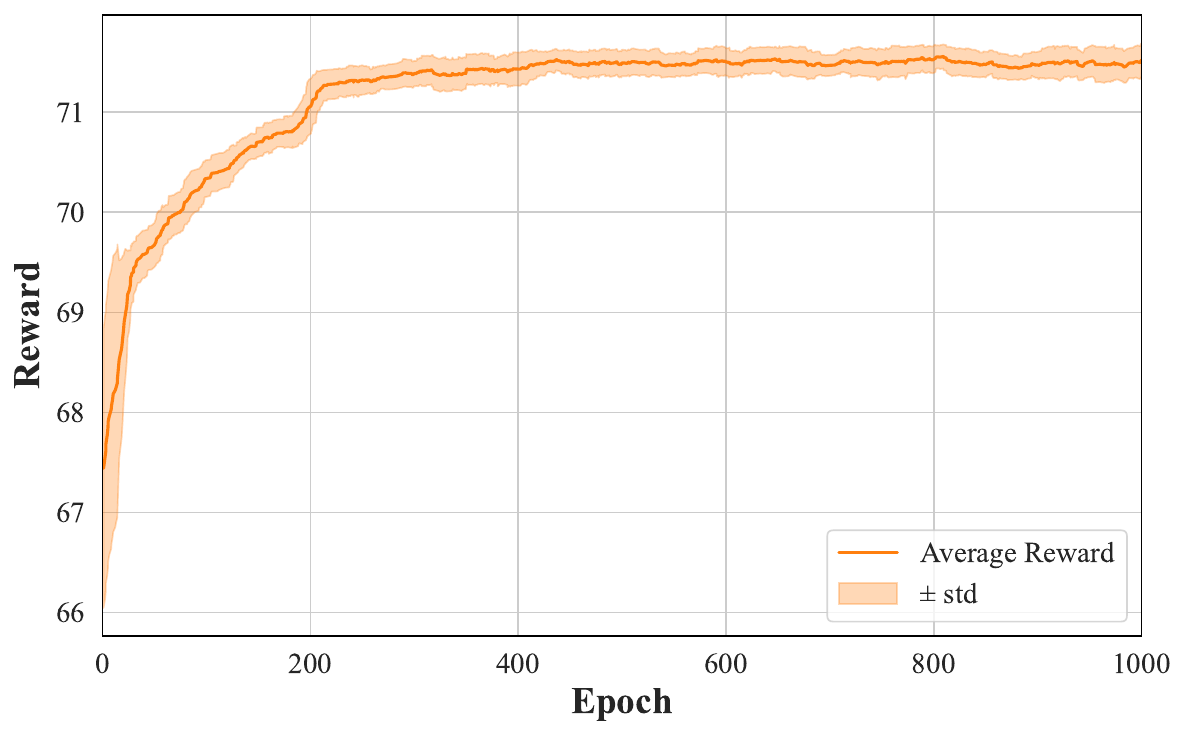}}& 
\subfloat[Interdiction agent of MCLIP20.\vspace{-2mm}]{\includegraphics[width=.49\linewidth]{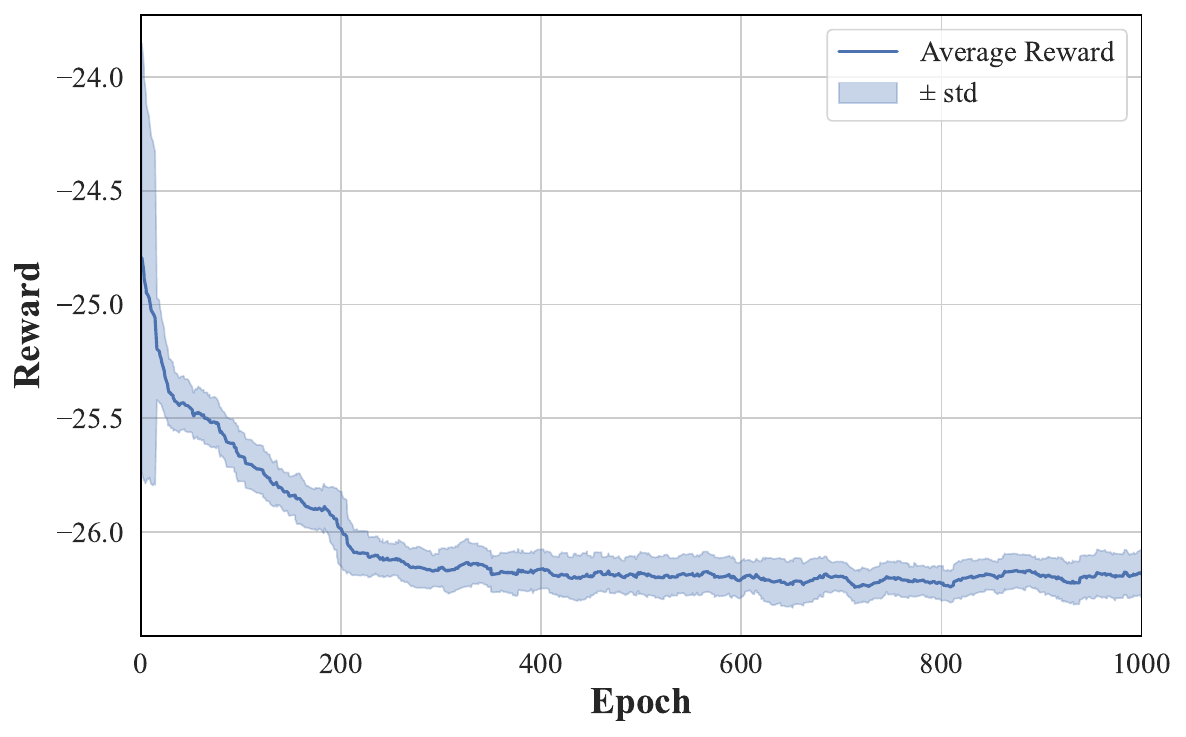}} &\\
\subfloat[Location agent of MCLIP100.\vspace{-2mm}]{\includegraphics[width=.49\linewidth]{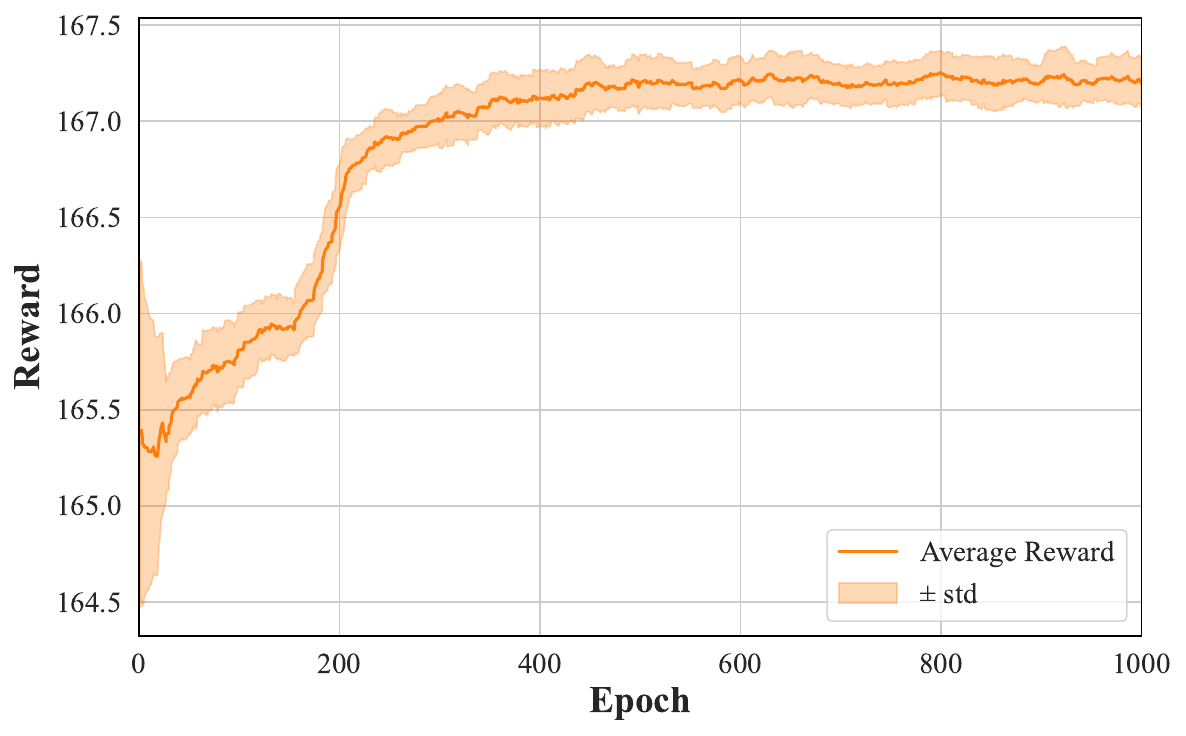}} &
\subfloat[Interdiction agent of MCLIP100.\vspace{-2mm}]{\includegraphics[width=.49\linewidth]{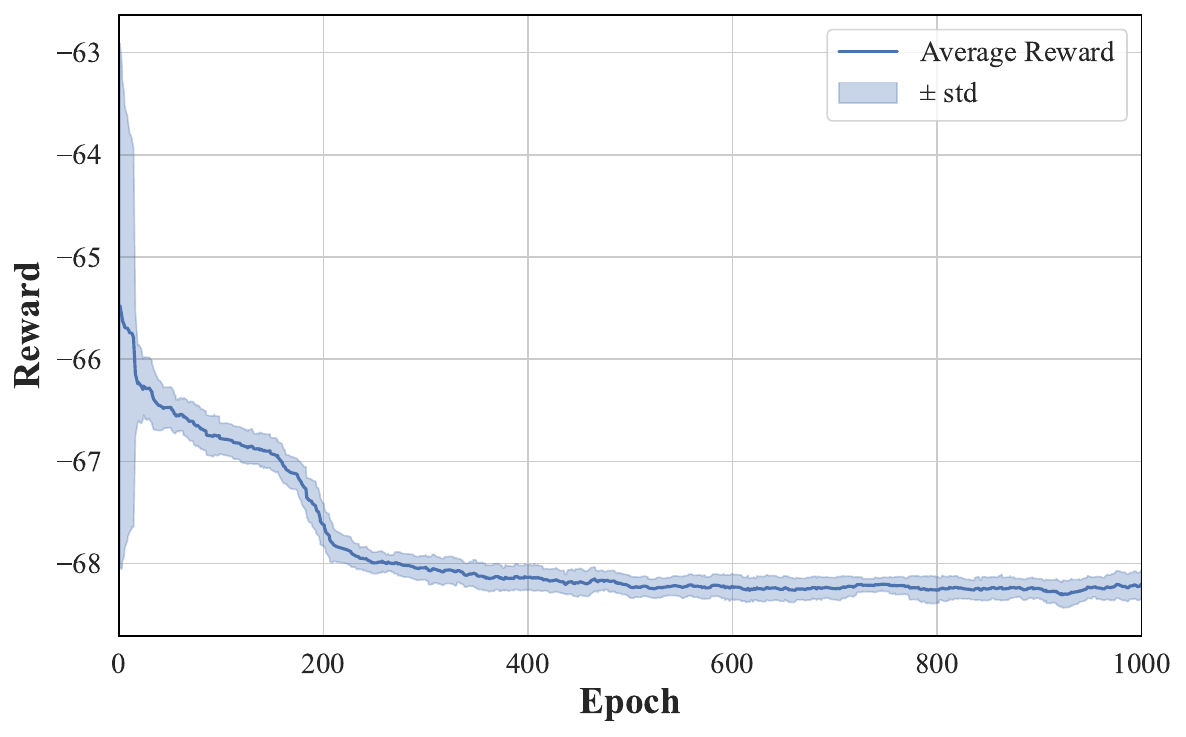}}
\end{tabular}
\caption{Training curves of dual agents across various scales.}
\label{fig:training}
\end{figure}

Additionally, we provide the training curves of dual agents across various scales in Fig. \ref{fig:training}. We can observe that initial phases exhibit high variance due to exploration, a trend that becomes more pronounced in larger scales given the expanded search space. However, as training progresses, the fluctuations gradually subside and the agents consistently converge to a stable equilibrium, demonstrating the robustness and stability of our framework.

To ensure a fair comparison with the exact solver and heuristic methods, all instances for are solved individually instead of in parallel. We evaluate our DADRL under both greedy and sampling decoding strategies. Notably, the sampling decoding is augmented by our proposed Surrogate-based Ensemble Inference Strategy. In this setup, the ensemble size $E$ is fixed at 10, while sampling sizes $K$ of 128 and 1,280 are tested for comparative analysis. All experiments are conducted on the server with a GTX 4090 GPU and Intel(R) Xeon(R) Gold 6230 CPU @ 2.10GHz.

\begin{table*}[!t]
\caption{Comparison Results on Synthetic Datasets across Various Problem Scales.}
\label{tab:main_results} 
\resizebox{\textwidth}{!}{
\begin{threeparttable}
\begin{tabular}{cc|ccccc|ccccc|ccccc}
\toprule
\multicolumn{2}{c}{\multirow{2}{*}{}}           & \multicolumn{5}{c}{MCLIP20}                                                                & \multicolumn{5}{c}{MCLIP50}                                                                & \multicolumn{5}{c}{MCLIP100}                                                                \\
\multicolumn{2}{c}{}                            & Pre.              & Post.             & Obj.             & Gap (\%)        & Time (s)        & Pre.              & Post.             & Obj.             & Gap (\%)        & Time (s)        & Pre.              & Post.             & Obj.              & Gap (\%)        & Time (s)        \\ \midrule
\multirow{2}{*}{\rotatebox[origin=c]{90}{Exact}}         & Gurobi        & 18.6730          & 13.8310          & 32.5040          & 0.00\%          & 0.0790          & 46.6170          & 27.2910          & 73.9080          & 0.00\%          & 124.8021        & -                & -                & -                 & -               & 3600.0000$^\dagger$       \\
                                & Sequential    & 18.9630          & 12.2500          & 31.2130          & 3.97\%          & 0.5033          & 47.4140          & 23.7560          & 71.1700          & 3.70\%          & 0.6069          & 100.0000         & 55.3410          & 155.3410          & 7.61\%          & 0.7630          \\ \midrule
\multirow{5}{*}{\rotatebox[origin=c]{90}{Heuristics}}    & GM            & 18.2770          & 13.3770          & 31.6540          & 2.62\%          & 0.0218          & 45.0330          & 25.4640          & 70.4970          & 4.62\%          & 0.2707          & 99.1360          & 63.4120          & 162.5480          & 3.33\%          & 3.2197          \\
                                & GE            & 18.3900          & 13.6550          & 32.0450          & 1.41\%          & 0.0838          & 45.3220          & 26.0570          & 71.3790          & 3.42\%          & 3.1820          & 99.1140          & 63.9740          & 163.0880          & 3.01\%          & 66.0015         \\
                                & GI            & 18.4090          & 13.6090          & 32.0180          & 1.50\%          & 0.0515          & 45.2200          & 26.0570          & 71.2770          & 3.56\%          & 0.8160          & 99.1080          & 63.9130          & 163.0210          & 3.05\%          & 9.0720          \\
                                & Stingy        & 17.3040          & 12.4740          & 29.7780          & 8.39\%          & 0.0293          & 42.5210          & 23.8170          & 66.3380          & 10.24\%         & 1.2077          & 96.8970          & 61.8280          & 158.7250          & 5.60\%          & 29.7060         \\
                                & AS            & 14.9260          & 10.6590          & 25.5850          & 21.29\%         & 0.0067          & 35.2850          & 18.3680          & 53.6530          & 27.41\%         & 0.0552          & 86.7290          & 50.6900          & 137.4190          & 18.27\%         & 0.4363          \\ \midrule
\multirow{4}{*}{\rotatebox[origin=c]{90}{Meta-H}} & GA            & 18.3010          & 13.5460          & 31.8470          & 2.02\%          & 0.4246          & 43.2170          & 24.5180          & 67.7350          & 8.35\%          & 3.8069          & 96.8000          & 62.1260          & 158.9260          & 5.48\%          & 56.2154         \\
                                & SA            & 18.5600          & 13.7510          & 32.3110          & 0.59\%          & 0.1711          & 44.0150          & 25.1200          & 69.1350          & 6.46\%          & 0.6016          & 97.3430          & 62.6690          & 160.0120          & 4.84\%          & 1.9973          \\
                                & TS            & 18.5650          & 13.7830          & 32.3480          & 0.48\%          & 0.3869          & 45.7110          & 26.3010          & 72.0120          & 2.57\%          & 10.0686         & 99.3710          & 64.1520          & 163.5230          & 2.75\%          & 151.6598        \\
                                & VNS           & 18.4960          & 13.7630          & 32.2590          & 0.75\%          & 0.8993          & 43.6070          & 24.7080          & 68.3150          & 7.57\%          & 8.7535          & 96.8670          & 62.3040          & 159.1710          & 5.34\%          & 46.6895         \\ \midrule
\multirow{6}{*}{\rotatebox[origin=c]{90}{Learning}}       & DADRL-AM-Greedy  & 18.5110          & 13.5130          & 32.0240          & 1.48\%          & 0.0110          & 45.2950          & 25.4300          & 70.7250          & 4.31\%          & 0.0169          & 99.4140          & 66.9110          & 166.3250          & 1.08\%          & 0.0257          \\
                                & DADRL-AM-128     & 18.5950          & 13.7010          & 32.2960          & 0.64\%          & 0.0124          & 45.8940          & 26.2460          & 72.1400          & 2.39\%          & 0.0200          & 99.6280          & 67.7100          & 167.3380          & 0.48\%          & 0.0330          \\
                                & DADRL-AM-1280    & 18.6050          & 13.7230          & 32.3280          & 0.54\%          & 0.0208          & 45.9940          & 26.4020          & 72.3960          & 2.05\%          & 0.0460          & 99.6640          & 67.7350          & 167.3990          & 0.44\%          & 0.1020          \\ \cmidrule{2-17} 
                                & DADRL-ADNet-Greedy & 18.5030          & 13.5420          & 32.0450          & 1.41\%          & 0.0136          & 45.5700          & 25.8820          & 71.4520          & 3.32\%          & 0.0217          & 99.2960          & 67.1550          & 166.4510          & 1.01\%          & 0.0325          \\
                                & DADRL-ADNet-128    & 18.5710          & 13.7300          & 32.3010          & 0.62\%          & 0.0151          & 46.1050          & 26.6310          & 72.7360          & 1.59\%          & 0.0247          & 99.6490          & 68.3440          & 167.9930          & 0.09\%          & 0.0415          \\
                                & DADRL-ADNet-1280   & \textbf{18.5890} & \textbf{13.7710} & \textbf{32.3600} & \textbf{0.44\%} & \textbf{0.0284} & \textbf{46.2200} & \textbf{26.7470} & \textbf{72.9670} & \textbf{1.27\%} & \textbf{0.0690} & \textbf{99.7060} & \textbf{68.4390} & \textbf{168.1450} & \textbf{0.00\%} & \textbf{0.1649} \\ \bottomrule
\end{tabular}
\begin{tablenotes} 
\footnotesize
\item $^*$ `Pre.' and `Post.' denote pre- and post-interdiction coverage, respectively. `Obj.' represents the objective function of MCLIP, defined as the sum of `Pre.' and `Post.'. The best-performing method (highest objective value) is highlighted in \textbf{bold}. All results are reported as the average of 1,000 independently generated instances for each problem scale.
\item $^\dagger$ Gurobi fails to find a feasible solution for MCLIP100 within the maximum time limit of 1 hour.
\end{tablenotes} 
\end{threeparttable} 
}
\end{table*}

\subsubsection{Baseline Algorithms}

To comprehensively evaluate the performance of DADRL, we compare it against three categories of algorithms: Exact Methods, Heuristics, and Metaheuristics. Consistent with the prevalent nested optimization framework in bilevel programming, our comparative study focuses on the selection of the upper-level location strategy. For all baseline algorithms (except the exact solver), the lower-level interdiction problem is approximated using a standardized Greedy Interdiction Heuristic. The heuristic iteratively removes the facility that contributes most to the current coverage until $r$ facilities are interdicted, serving as a computationally efficient surrogate for the follower's response. Noted that we set the maximum time limit of 1 hour for all baselines, and the baselines can be categorized as follows:

\textbf{(1) Exact Methods.} We employ the commercial solver Gurobi \cite{gurobi} to solve the equivalent single-level MIP formulation [MCLIP-2], which provides the theoretical optimal bound for small-scale instances. Additionally, to demonstrate the necessity of MCLIP, we include a baseline denoted as \textit{Sequential}. Specifically, \textit{Sequential} first solves a standard MCLP ignoring potential interdiction, and subsequently evaluates its performance under the worst-case interdiction.

\textbf{(2) Heuristics.} We implement a suite of classical heuristics specifically tailored for facility location problems. These include constructive methods such as Greedy Myopic (GM) \cite{GM}, which maximizes marginal gain sequentially, and Stingy \cite{Stingy}, a reverse greedy approach. We also include improvement-based methods: Greedy Exchange (GE) \cite{GE} and its more aggressive variant Global Interchange (GI) \cite{GI}, which optimize solutions via facility swapping, as well as Alternate Selection (AS) \cite{AS} which combines addition and substitution strategies.

\textbf{(3) Metaheuristics.} We also adapt several widely-used metaheuristics to the nested MCLIP structure. The comparison includes both population-based and improvement-based evolutionary approaches: Genetic Algorithm (GA) \cite{GA}, Simulated Annealing (SA) \cite{SA}, Tabu Search (TS) \cite{TS}, and Variable Neighborhood Search (VNS) \cite{VNS}.

\begin{table*}[!t]
\caption{Comparison Results on Cross-Scale instances.}
\label{tab:scale_results} 
\resizebox{\textwidth}{!}{
\begin{threeparttable}
\begin{tabular}{cc|ccccc|ccccc|ccccc|ccccc}
\toprule
\multicolumn{2}{c}{\multirow{2}{*}{}}             & \multicolumn{5}{c}{MCLIP200*}                                                                  & \multicolumn{5}{c}{MCLIP300*}                                                                  & \multicolumn{5}{c}{MCLIP400*}                                                                  & \multicolumn{5}{c}{MCLIP500*}                                                                  \\
\multicolumn{2}{c}{}                              & Pre.               & Post.              & Obj.              & Gap (\%)        & Time (s)        & Pre.               & Post.              & Obj.              & Gap (\%)        & Time (s)        & Pre.               & Post.              & Obj.              & Gap (\%)        & Time (s)        & Pre.               & Post.              & Obj.              & Gap (\%)        & Time (s)        \\ \midrule
\multirow{2}{*}{\rotatebox[origin=c]{90}{Exact}}          & Gurobi          & -                 & -                 & -                 & -               & 3600.0000$^\dagger$       & -                 & -                 & -                 & -               & 3600.0000$^\dagger$       & -                 & -                 & -                 & -               & 3600.0000$^\dagger$       & -                 & -                 & -                 & -               & 3600.0000$^\dagger$       \\
                                & Sequential      & 200.0000          & 116.4900          & 316.4900          & 6.89\%          & 0.9608          & 300.0000          & 179.6200          & 479.6200          & 6.32\%          & 1.2828          & 400.0000          & 244.1300          & 644.1300          & 5.83\%          & 1.5577          & 500.0000          & 310.9600          & 810.9600          & 5.30\%          & 1.8472          \\ \midrule
\multirow{5}{*}{\rotatebox[origin=c]{90}{Heuristics}}     & GM              & 197.4200          & 130.1300          & 327.5500          & 3.63\%          & 8.5940          & 295.2800          & 196.9100          & 492.1900          & 3.86\%          & 16.5255         & 393.1000          & 264.2000          & 657.3000          & 3.91\%          & 26.6723         & 492.2000          & 331.7200          & 823.9200          & 3.78\%          & 39.4051         \\
                                & GE              & 198.4700          & 131.8100          & 330.2800          & 2.83\%          & 212.7941        & 297.8600          & 199.4500          & 497.3100          & 2.86\%          & 499.3638        & 397.1400          & 265.9500          & 663.0900          & 3.06\%          & 895.9174        & 497.0000          & 333.5500          & 830.5500          & 3.01\%          & 1467.1463       \\
                                & GI              & 198.2400          & 131.1100          & 329.3500          & 3.10\%          & 30.9542         & 297.9800          & 199.6900          & 497.6700          & 2.79\%          & 70.0824         & 397.0900          & 266.9500          & 664.0400          & 2.92\%          & 125.1986        & 497.1900          & 335.0700          & 832.2600          & 2.81\%          & 202.9211        \\
                                & Stingy          & 191.6800          & 124.7600          & 316.4400          & 6.90\%          & 44.1356         & 287.8500          & 187.8800          & 475.7300          & 7.08\%          & 62.0670         & 382.2000          & 250.4600          & 632.6600          & 7.51\%          & 81.0477         & 477.2200          & 314.7200          & 791.9400          & 7.52\%          & 104.0419        \\
                                & AS              & 170.8800          & 100.6200          & 271.5000          & 20.12\%         & 0.8518          & 252.7100          & 149.9300          & 402.6400          & 21.35\%         & 1.4062          & 339.2000          & 203.5700          & 542.7700          & 20.65\%         & 2.0208          & 421.8400          & 252.9300          & 674.7700          & 21.20\%         & 2.5304          \\ \midrule
\multirow{4}{*}{\rotatebox[origin=c]{90}{Meta-H}} & GA              & 192.1400          & 124.7800          & 316.9200          & 6.76\%          & 79.9213         & 286.3000          & 187.7800          & 474.0800          & 7.40\%          & 100.4678        & 382.3900          & 252.3600          & 634.7500          & 7.20\%          & 126.0253        & 476.8400          & 315.5900          & 792.4300          & 7.46\%          & 152.0615        \\
                                & SA              & 192.9400          & 126.6600          & 319.6000          & 5.97\%          & 2.6644          & 287.9400          & 191.6500          & 479.5900          & 6.32\%          & 3.2673          & 384.1400          & 254.1500          & 638.2900          & 6.69\%          & 3.9258          & 480.1600          & 320.0500          & 800.2100          & 6.55\%          & 4.5044          \\
                                & TS              & 198.9400          & 132.2600          & 331.2000          & 2.56\%          & 467.0116        & 298.3800          & 199.4100          & 497.7900          & 2.77\%          & 875.5914        & 397.2800          & 266.1400          & 663.4200          & 3.01\%          & 1484.6208       & 497.2300          & 335.1200          & 832.3500          & 2.80\%          & 2136.1927       \\
                                & VNS             & 189.3100          & 123.0300          & 312.3400          & 8.11\%          & 52.9925         & 279.1100          & 180.9600          & 460.0700          & 10.14\%         & 60.1990         & 369.4200          & 238.3900          & 607.8100          & 11.14\%         & 66.5165         & 454.2100          & 288.7300          & 742.9400          & 13.24\%         & 76.4778         \\ \midrule
\multirow{6}{*}{\rotatebox[origin=c]{90}{Learning}}       & DADRL-AM-Greedy    & 199.5000          & 138.5100          & 338.0100          & 0.56\%          & 0.0319          & 299.4800          & 210.4300          & 509.9100          & 0.40\%          & 0.0316          & 399.4900          & 282.5400          & 682.0300          & 0.29\%          & 0.0325          & 499.1800          & 354.0600          & 853.2400          & 0.36\%          & 0.0314          \\
                                & DADRL-AM-128       & 199.7600          & 139.8800          & 339.6400          & 0.08\%          & 0.0492          & 299.7100          & 211.9600          & 511.6700          & 0.06\%          & 0.0610          & 399.5800          & 284.0200          & 683.6000          & 0.06\%          & 0.0747          & \textbf{499.5100} & \textbf{356.8100} & \textbf{856.3200} & \textbf{0.00\%} & \textbf{0.0915} \\
                                & DADRL-AM-1280      & 199.7800          & 140.1000          & 339.8800          & 0.01\%          & 0.2672          & 299.7800          & 211.9200          & 511.7000          & 0.05\%          & 0.4105          & \textbf{399.7200} & \textbf{284.3000} & \textbf{684.0200} & \textbf{0.00\%} & \textbf{0.5873} & 499.6100          & 356.5200          & 856.1300          & 0.02\%          & 0.6983          \\ \cmidrule{2-22} 
                                & DADRL-ADNet-Greedy & 199.2300          & 138.2300          & 337.4600          & 0.72\%          & 0.0378          & 299.2800          & 210.2100          & 509.4900          & 0.48\%          & 0.0380          & 399.3200          & 281.4700          & 680.7900          & 0.47\%          & 0.0372          & 498.9300          & 352.8300          & 851.7600          & 0.53\%          & 0.0395          \\
                                & DADRL-ADNet-128    & 199.7000          & 140.0900          & 339.7900          & 0.03\%          & 0.0620          & 299.6900          & 212.2200          & 511.9100          & 0.01\%          & 0.0784          & 399.5400          & 284.3800          & 683.9200          & 0.01\%          & 0.0905          & 499.4300          & 356.2400          & 855.6700          & 0.08\%          & 0.1136          \\
                                & DADRL-ADNet-1280   & \textbf{199.6200} & \textbf{140.2800} & \textbf{339.9000} & \textbf{0.00\%} & \textbf{0.3010} & \textbf{299.6400} & \textbf{212.3200} & \textbf{511.9600} & \textbf{0.00\%} & \textbf{0.4681} & 399.5900          & 284.0300          & 683.6200          & 0.06\%          & 0.6362          & 499.3300          & 356.5000          & 855.8300          & 0.06\%          & 0.7973          \\ \bottomrule
\end{tabular}
\begin{tablenotes} 
\footnotesize
\item $^*$ `Pre.' and `Post.' denote pre- and post-interdiction coverage, respectively. `Obj.' represents the objective function of MCLIP, defined as the sum of `Pre.' and `Post.'. The best-performing method (highest objective value) is highlighted in \textbf{bold}. All results are reported as the average of 100 independently generated instances for each problem scale.
\item $^\dagger$ Gurobi fails to find a feasible solution for MCLIP200, MCLIP300, MCLIP400, and MCLIP500 within the maximum time limit of 1 hour.\vspace{-10mm}
\end{tablenotes} 
\end{threeparttable} 
}
\end{table*}

\begin{figure*}[t]
\centering
\setlength{\tabcolsep}{1pt}
\begin{tabular}{ccc}
\subfloat[BJ671]{\includegraphics[width=.2\linewidth]{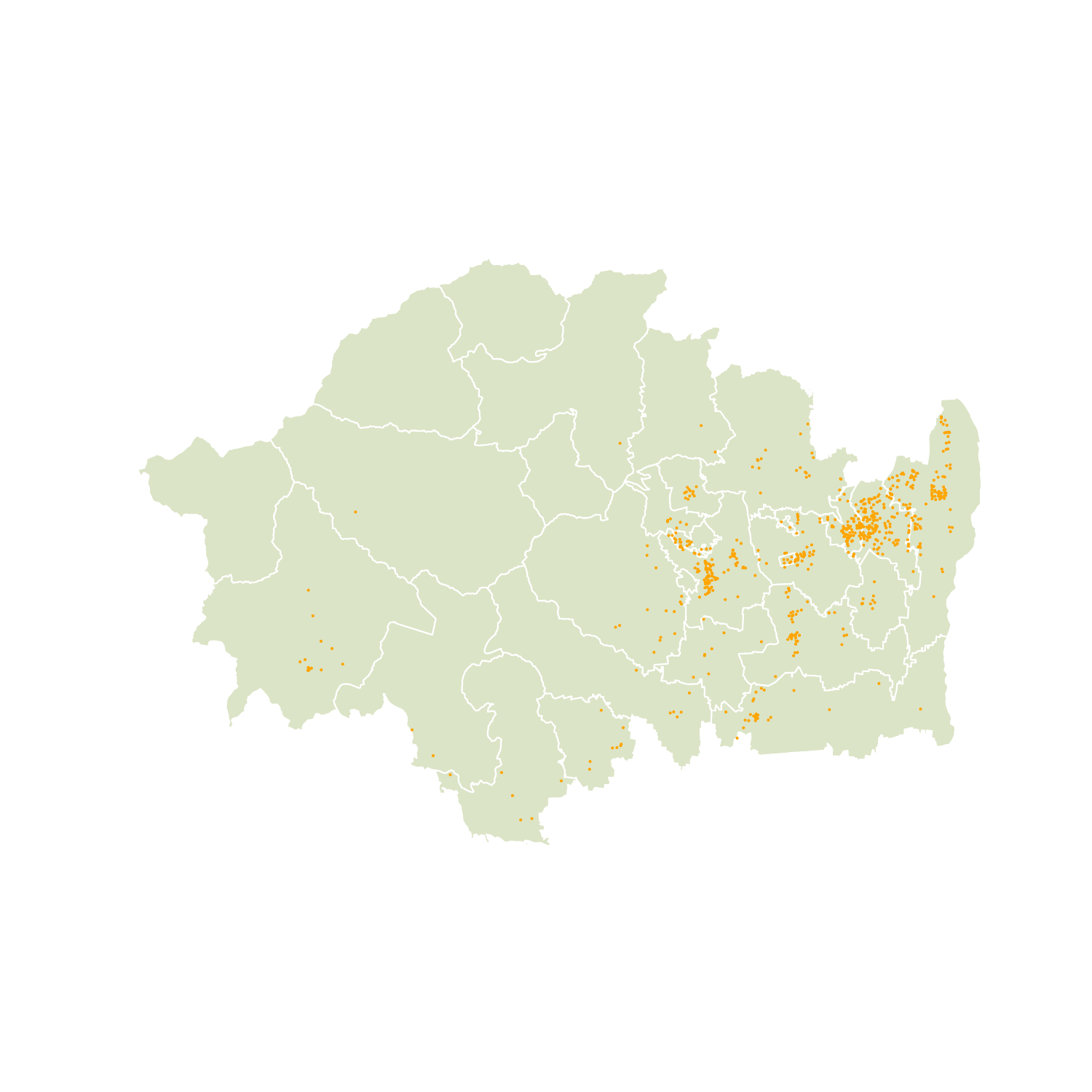}} 
\subfloat[BJ795]{\includegraphics[width=.2\linewidth]{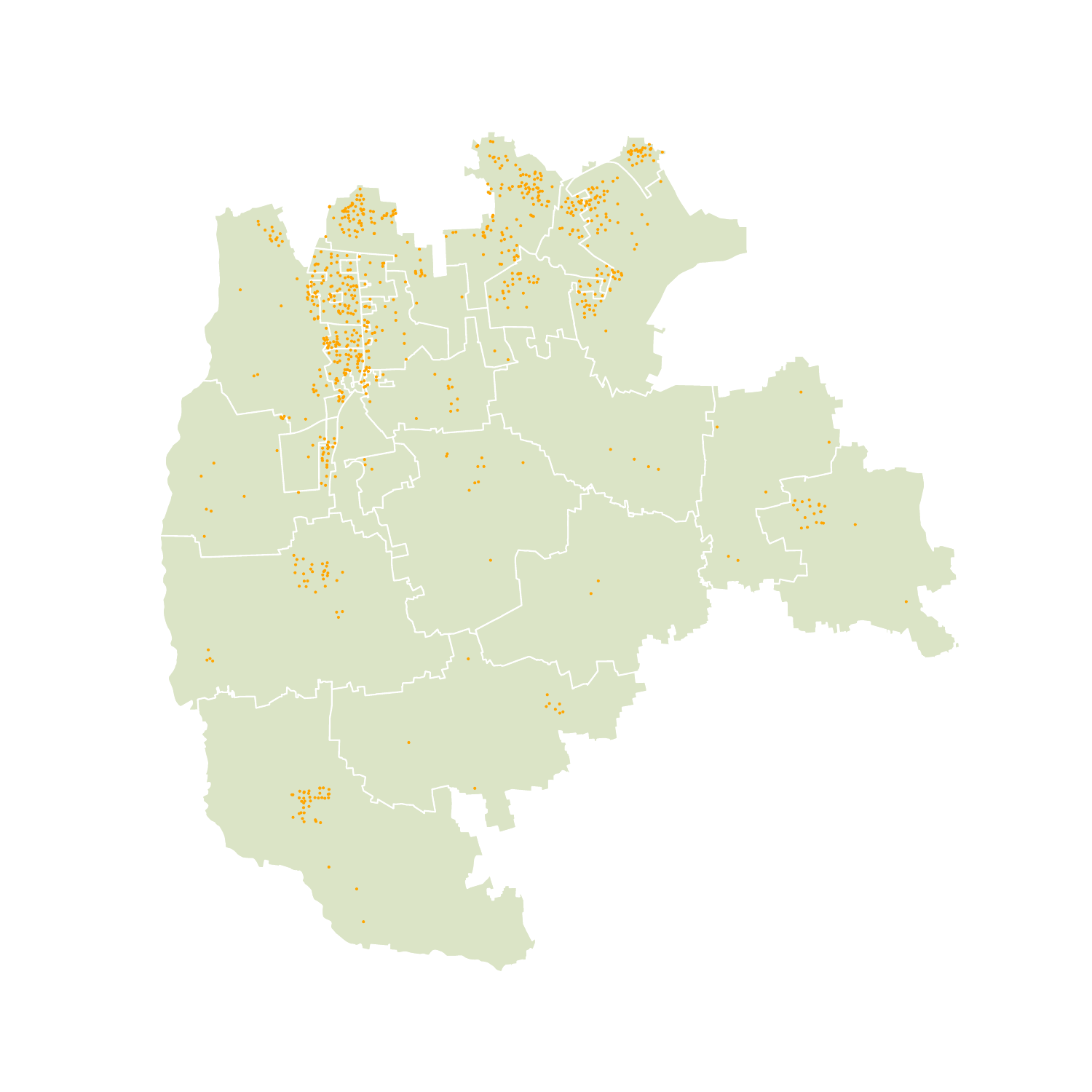}} 
\subfloat[BJ1202]{\includegraphics[width=.2\linewidth]{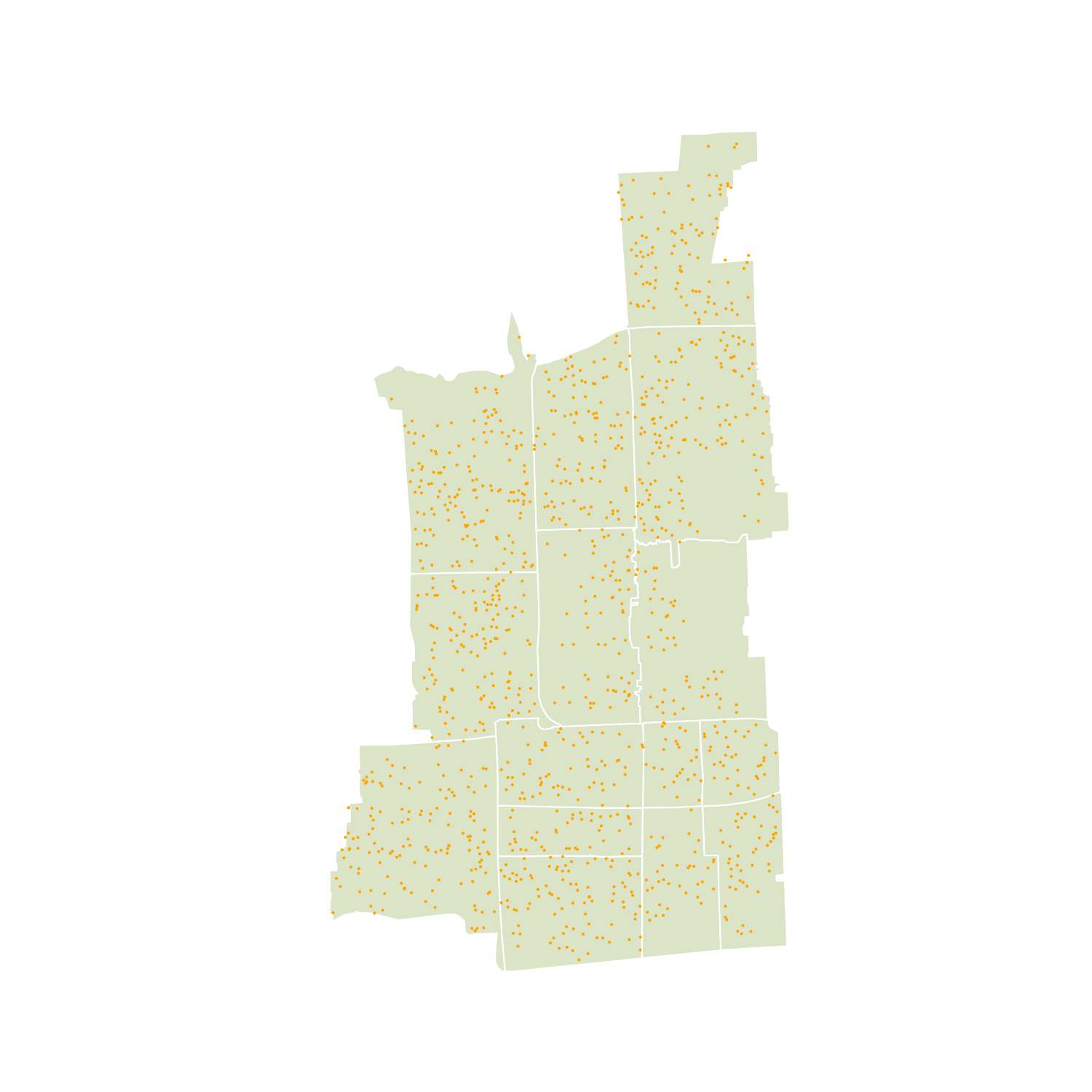}}  
\subfloat[BJ1416]{\includegraphics[width=.2\linewidth]{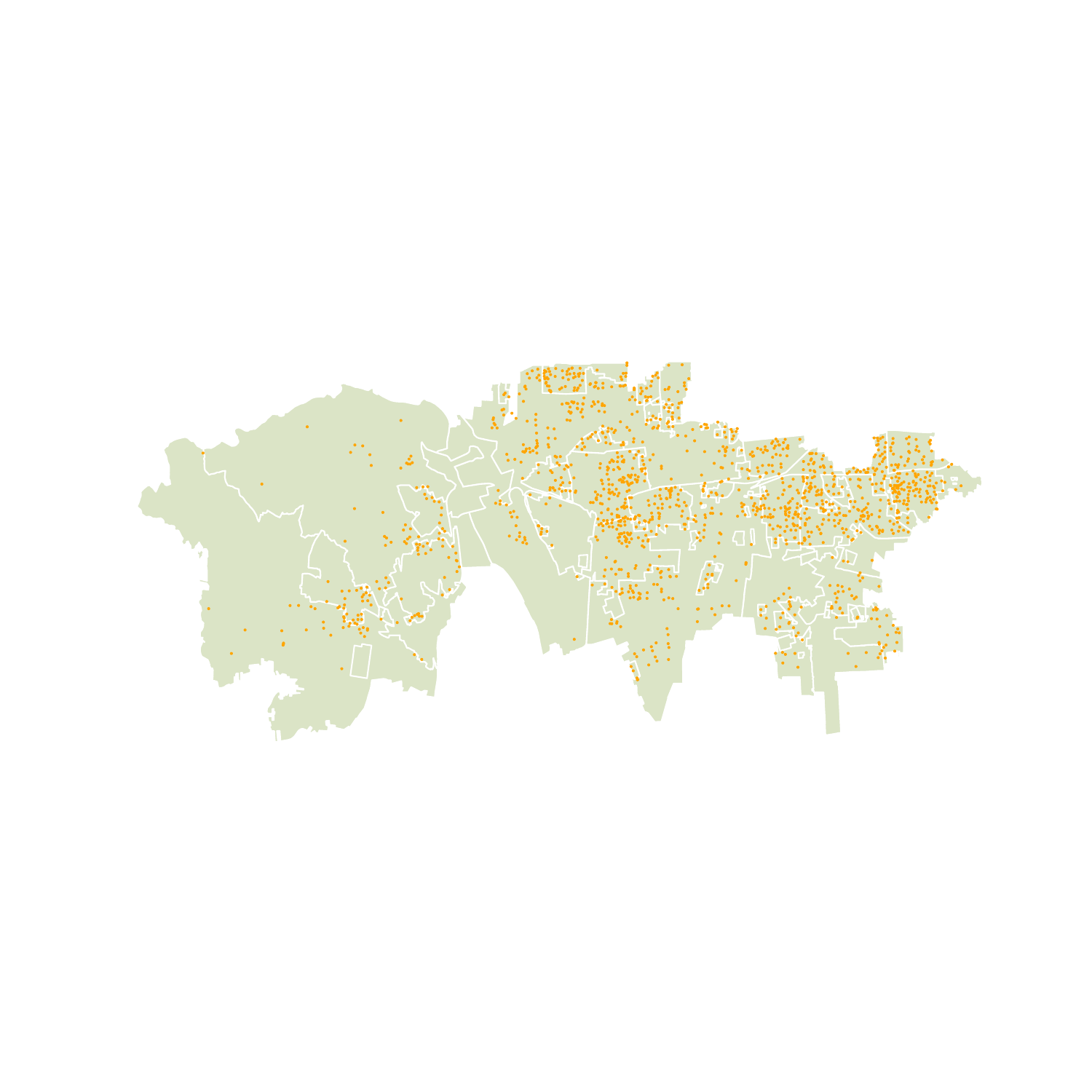}} 
\subfloat[BJ2043]{\includegraphics[width=.2\linewidth]{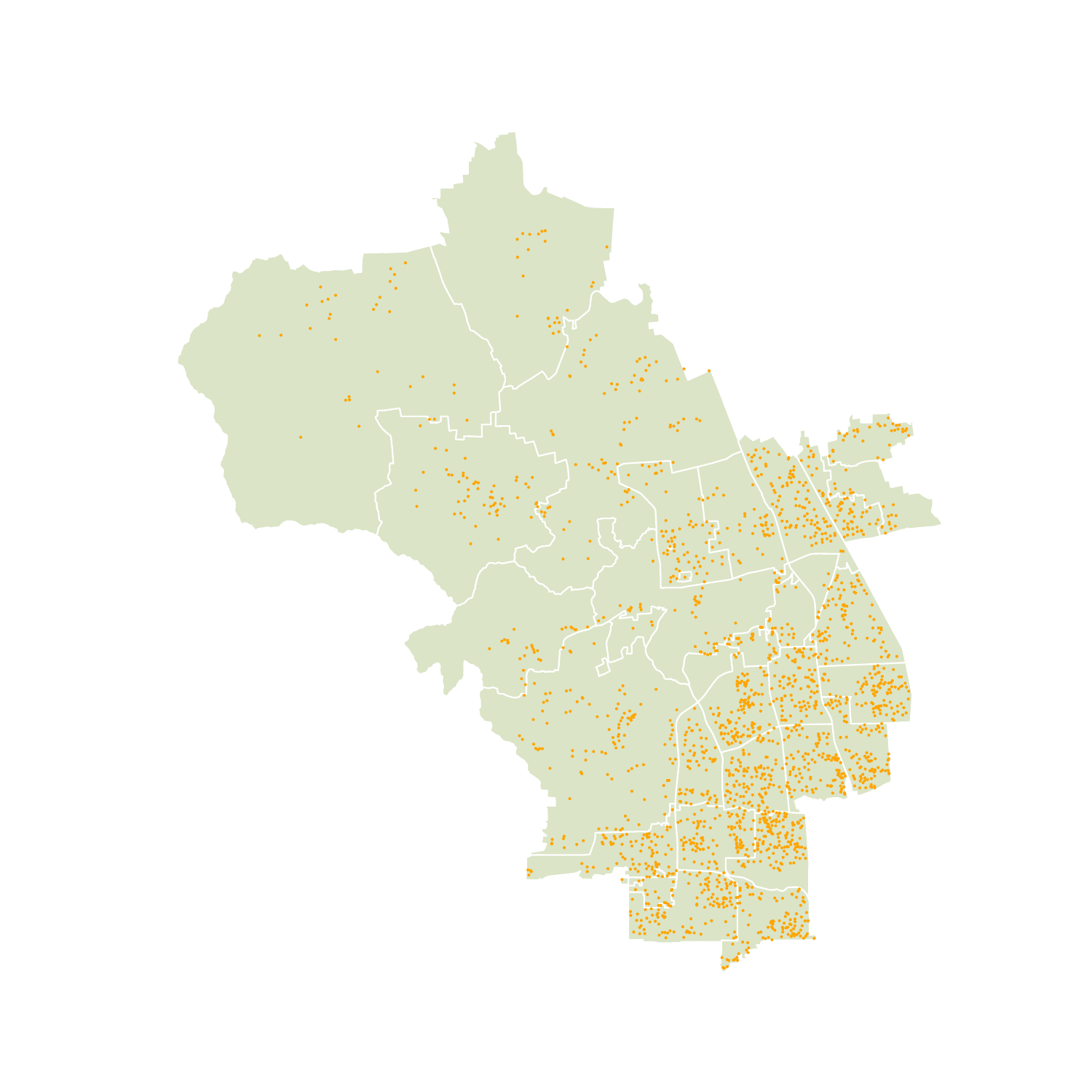}} 
\end{tabular}
\caption{Visualizations of selected instances from the Beijing dataset, where BJ$x$ represents an instance with $x$ nodes.}
\label{fig:beijing}
\end{figure*}

\subsection{Comparison Study on Synthetic Datasets}

Table \ref{tab:main_results} presents a comparative performance analysis of the proposed DADRL framework against baseline algorithms. We evaluate performance based on computational efficiency and solution quality, where the latter is measured by the objective value and the optimality gap. Specifically, the optimality gap is calculated against the best-performing method among all baselines, and we generate 1,000 independent instances for each problem scale.

\subsubsection{Comparison with Exact Methods}

Regarding exact methods, although Gurobi solves the small-scale MCLIP20 in 0.08s, its computational cost grows exponentially with problem scale. It requires approximately 2 minutes to resolve MCLIP50 and fails to provide even a feasible solution for MCLIP100 within the one hour. This scalability bottleneck is critical, as real-world scenarios typically involve hundreds or thousands of nodes. Compared to the Sequential baseline, the results demonstrate that ignoring potential interdiction leads to a severe degradation in post-interdiction coverage. In contrast, our DADRL not only maintains a minimal optimality gap on small-scale instances but also generates high-quality solutions for MCLIP100 within one second, demonstrating significant potential for real-world applications.

\subsubsection{Comparison with Heuristics}

Classical heuristics often struggle with local optima. Even the best-performing improvement heuristic GE maintains a 3.56\% gap on MCLIP50, while constructive methods like Stingy perform worse. Notably, we observe a significant reduction in the optimality gap on MCLIP100 compared to MCLIP50. This trend can be attributed to the specific experimental settings, particularly the variation in the number of facilities $p$. Regarding the heuristic baselines, GM and GI demonstrate the most favorable trade-off between solution quality and computational efficiency. Regarding DADRL, even its greedy decoding variant outperforms all baseline heuristics in terms of all metrics.

\subsubsection{Comparison with Meta-Heuristics}

Metaheuristics typically rely on extensive iterative optimization processes, which inherently result in higher computational overhead compared to simple heuristics. While methods like TS can yield competitive solutions, they incur high computational costs. In contrast, SA offers a moderate alternative, providing a reasonable trade-off between solution quality and runtime. Crucially, our DADRL surpasses even the best-performing metaheuristic TS in solution quality across all problem scales.

\subsubsection{Comparison between various backbones}

In the comparison between AM and ADNet, ADNet consistently achieves superior gaps across all scales. While ADNet incurs a slight computational cost due to the incorporation of specialized dynamic features for facility location, the performance gain makes the trade-off acceptable. Furthermore, regarding inference, our Surrogate-based Ensemble Inference Strategy provides a significant performance boost, which scales positively with the sampling size. Therefore, the competitive results achieved on both backbones demonstrate the robust effectiveness and broad versatility of the DADRL framework.

\subsection{Performance on Cross-Scale Instances}

To investigate the cross-scale scalability, we extend our evaluation to larger instances with sizes ranging from $N=200$ to $N=500$. We randomly generate a test set of 100 instances for each scale, and all dataset settings are consistent with the scale of 100. Meanwhile, we directly apply the DADRL model pre-trained on MCLIP100 to solve larger-scale instances, which can better demonstrate its generalization performance.

The results are shown in Table \ref{tab:scale_results}. As the problem scale expands, the superiority of DADRL over baseline methods becomes increasingly pronounced, particularly regarding computational efficiency. While exact solvers fail to provide feasible solutions and metaheuristics suffer from exponential runtime growth, our DADRL can generate superior solutions in less than 1 second.

Despite the overall dominance, we observe two situations among the results. First, regarding the backbone architecture, AM exhibits better performance than ADNet on MCLIP400 and MCLIP500. We speculate that ADNet's specialized modeling of dynamic features may limit its generalizability compared to the more generic structure of AM, while it is highly effective within the training distribution. Second, regarding the decoding strategy, increasing the sampling size from 128 to 1,280 yields slight degradation on MCLIP400 and MCLIP500, which is contrary to our intuition. This phenomenon can be attributed to the generalization boundary of the lower-level Interdiction Agent. As the model's performance naturally degrades on unseen scales, the reward estimation of surrogates may also be influenced. Therefore, a suboptimal solution may be considered as the best solution during the sampling decoding strategy. Nevertheless, DADRL consistently achieves the lowest optimality gaps across all scales, proving its robust generalization and effectiveness against all baselines.

\subsection{Performance on Real-world Datasets}

To validate the practical applicability of DADRL, we conduct extensive experiments on two real-world datasets: SJC and BJ, which have been mentioned before. Specifically, the BJ dataset contains 16 instances, each corresponding to a distinct administrative districts with graph sizes ranging from 169 to 2,472. Fig. \ref{fig:beijing} presents the visualization of several representative instances, highlighting the distinct distributional diversity inherent in these real-world scenarios. Considering the suboptimal performance of Stingy, AS, GA, and VNS observed in the synthetic experiments, we exclude these methods from the real-world evaluation. To ensure fairness, we directly apply the models pre-trained on MCLIP100 in real-world evaluation.

\begin{table*}[!t]
\caption{Comparison Results on Real-World Instances.}
\label{tab:real_world}
\resizebox{\textwidth}{!}{
\begin{threeparttable}
\begin{tabular}{c|cc|cc|cc|cc|cc|cc|cc|cc}
\toprule
\multirow{2}{*}{Instance} & \multicolumn{2}{c|}{Sequential} & \multicolumn{2}{c}{GM} & \multicolumn{2}{c}{GE} & \multicolumn{2}{c|}{GI} & \multicolumn{2}{c}{SA} & \multicolumn{2}{c|}{TS} & \multicolumn{2}{c}{DADRL-AM}       & \multicolumn{2}{c}{DADRL-ADNet}    \\
                           & Obj.          & Time (s)        & Obj.      & Time (s)    & Obj.    & Time (s)     & Obj.    & Time (s)     & Obj.      & Time (s)    & Obj.    & Time (s)      & Obj.           & Time (s)        & Obj.           & Time (s)        \\ \midrule
SJC324 & 489 & 1.6375 & 533 & 16.6469 & 545 & 396.4606 & 529 & 82.2518 & 532 & 2.8693 & 534 & 875.5275 & 560 & 0.5573 & \textbf{561} & \textbf{0.5430} \\
SJC402 & 621 & 1.5307 & 681 & 23.8142 & 672 & 710.2530 & 670 & 122.1702 & 675 & 3.9051 & 671 & 1285.8185 & 698 & 0.7120 & \textbf{700} & \textbf{0.5783} \\
SJC500 & 786 & 1.8049 & 858 & 34.6497 & 844 & 761.8493 & 850 & 162.9582 & 836 & 4.0827 & 845 & 1907.2770 & 886 & 0.7334 & \textbf{890} & \textbf{0.7590} \\
SJC708 & 1134 & 2.4732 & 1221 & 62.5221 & 1204 & 1614.8136 & 1184 & 222.9876 & 1111 & 4.9258 & 1189 & 3562.1723 & 1240 & 0.9491 & \textbf{1257} & \textbf{1.0415} \\
SJC818 & 1290 & 2.7019 & 1380 & 81.5548 & 1371 & 2596.0415 & 1329 & 565.7854 & 1347 & 6.3283 & 1375 & 3600.0000$^\dagger$ & 1402 & 1.1810 & \textbf{1412} & \textbf{1.2188} \\ \midrule
BJ169 & 193 & 0.8512 & 221 & 5.3723 & 228 & 70.6092 & 226 & 6.8353 & 225 & 1.2261 & 230 & 200.0898 & 288 & 0.3403 & \textbf{295} & \textbf{0.3519} \\
BJ174 & 224 & 0.8715 & 294 & 5.6053 & 306 & 83.1462 & 314 & 17.5824 & 265 & 1.0047 & 309 & 208.5393 & 313 & 0.3703 & \textbf{317} & \textbf{0.3797} \\
BJ266 & 288 & 1.1141 & 334 & 9.4650 & 391 & 130.9529 & 501 & 18.8522 & 498 & 1.2828 & 324 & 372.0119 & \textbf{508} & \textbf{0.4028} & 503 & 0.4102 \\
BJ273 & 344 & 1.1289 & 332 & 9.8445 & 445 & 136.1378 & 409 & 15.3295 & 485 & 0.9614 & 456 & 379.8689 & 490 & 0.4347 & \textbf{509} & \textbf{0.4228} \\
BJ278 & 305 & 1.1578 & 324 & 9.9627 & \textbf{535} & \textbf{217.3358} & 355 & 21.8248 & 531 & 1.1717 & 353 & 396.5568 & 532 & 0.4294 & 523 & 0.4567 \\
BJ346 & 468 & 1.3621 & \textbf{627} & \textbf{11.6335} & 544 & 335.2005 & 611 & 41.4075 & 592 & 1.5647 & 529 & 541.8352 & 600 & 0.5289 & 621 & 0.5693 \\
BJ522 & 716 & 1.8248 & 850 & 20.2250 & 863 & 476.9627 & 853 & 104.6807 & 858 & 2.0261 & 767 & 1021.3858 & 829 & 0.8204 & \textbf{919} & \textbf{0.8478} \\
BJ671 & 855 & 2.3044 & 942 & 33.8551 & 971 & 988.5391 & 1072 & 115.4143 & 1018 & 2.1317 & 1091 & 1504.4434 & 1049 & 1.0630 & \textbf{1224} & \textbf{1.0837} \\
BJ795 & 1112 & 2.7191 & 1216 & 42.6603 & 1207 & 1100.0686 & 1252 & 72.1138 & 1216 & 2.3843 & 1219 & 2003.3792 & 1272 & 1.2189 & \textbf{1351} & \textbf{1.2094} \\
BJ796 & 1254 & 2.6414 & 1370 & 45.3755 & 1338 & 2326.5897 & 1367 & 372.1507 & 1371 & 2.9990 & 1334 & 2085.1801 & \textbf{1417} & \textbf{1.4594} & 1412 & 1.4535 \\
BJ910 & 1103 & 3.3892 & 1157 & 46.1088 & 1223 & 1376.1662 & 1307 & 218.5050 & 1615 & 2.2663 & 1196 & 2422.4514 & 1617 & 1.3842 & \textbf{1633} & \textbf{1.4527} \\
BJ946 & 1399 & 3.1765 & 1460 & 57.3056 & 1454 & 1550.6655 & 1557 & 372.4038 & 1535 & 3.3234 & 1525 & 2733.6205 & \textbf{1732} & \textbf{1.3473} & 1621 & 1.4829 \\
BJ1202 & 1936 & 3.8122 & 2023 & 85.3466 & 2024 & 2724.3495 & 2008 & 348.5137 & 1978 & 3.4842 & 1961 & 3600.0000$^\dagger$ & 2086 & 1.6748 & \textbf{2113} & \textbf{1.7834} \\
BJ1416 & 2241 & 4.7790 & 1944 & 111.8829 & 2232 & 2836.0955 & 2174 & 1035.5209 & 2305 & 3.5645 & 2294 & 3600.0000$^\dagger$ & 2475 & 0.6202 & \textbf{2484} & \textbf{2.1700} \\
BJ2043 & 2833 & 8.3740 & 3169 & 205.5683 & 3010 & 3600.0000$^\dagger$ & 3274 & 820.3374 & 3451 & 4.6267 & 3473 & 3600.0000$^\dagger$ & 3103 & 3.3845 & \textbf{3679} & \textbf{3.3013} \\
BJ2472 & 3785 & 9.3307 & 3885 & 284.1374 & 3999 & 3600.0000$^\dagger$ & 3886 & 2910.8659 & 4206 & 6.4199 & 4177 & 3600.0000$^\dagger$ & \textbf{4378} & \textbf{3.5793} & 4310 & 3.6783 \\ \bottomrule
\end{tabular}
\begin{tablenotes} 
\footnotesize
\item $^*$ `Obj.' represents the objective function of MCLIP, defined as the sum of `Pre.' and `Post.'. The best-performing method (highest objective value) is highlighted in \textbf{bold}. By default, the decoding strategy for all DADRL algorithms employs our proposed Surrogate-based Ensemble Inference Strategy.
\item $^\dagger$ We impose a maximum time limit of 3,600 seconds per instance for all baseline methods.
\end{tablenotes} 
\end{threeparttable}
}
\end{table*}

Table \ref{tab:real_world} reports the comparative results on 21 real-world instances derived from SJC and BJ. Our proposed DADRL framework exhibits dominance in solution quality, achieving the best objective values in 19 out of 21 instances. Specifically, ADNet secures the best known solutions in most scenarios, while AM also leads in instances like BJ266 and BJ2472. Although classical heuristics (GE and GM) yield slightly higher objective values in the remaining two instances, the performance gap is negligible. However, this marginal difference in quality comes at a high computational cost for the baseline methods. As illustrated in the table, baseline methods such as GE and TS suffer from severe scalability issues, hitting the maximum time limit (3,600s) on instances with over 1,000 nodes. In contrast, our DADRL consistently generates high-quality solutions with remarkable efficiency, requiring no more than 4 seconds even for the largest instance. These results conclusively demonstrate the effectiveness and practicality of DADRL in solving large-scale real-world problems.

\subsection{Ablation Study}

We conduct an ablation study to isolate and verify the contributions of our framework's components. Specifically, DADRL comprises two integral parts: the adversarial training pipeline and the surrogate-based ensemble inference strategy. Since the training pipeline is fundamental to addressing the hierarchical bi-level structure of MCLIP, our ablation focuses exclusively on the inference mechanism. 

\begin{figure}[t]
\centering
\setlength{\tabcolsep}{1pt}
\begin{tabular}{c}
\subfloat[Ablation study based on the backbone of AM.\vspace{-2mm}]{\includegraphics[width=\linewidth]{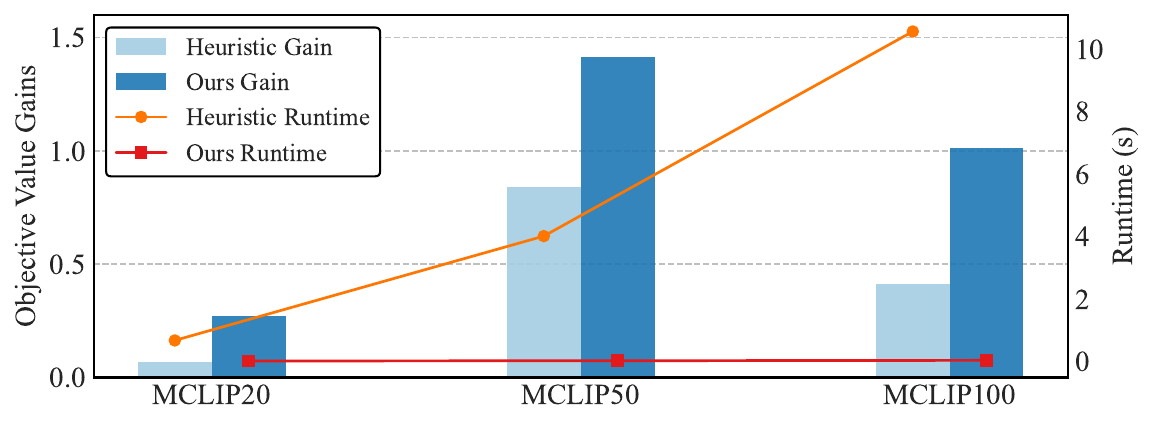}}\\
\subfloat[Ablation study based on the backbone of ADNet.\vspace{-2mm}]{\includegraphics[width=\linewidth]{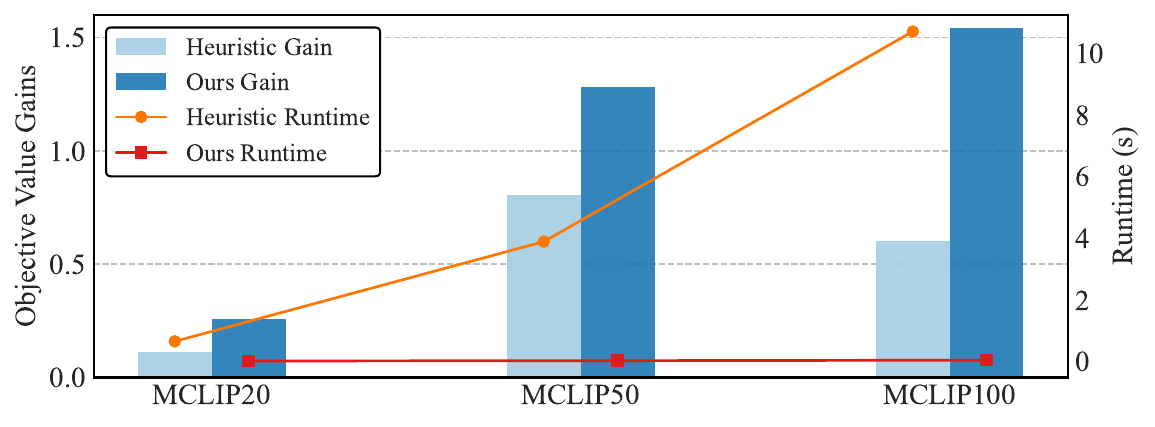}}
\end{tabular}
\caption{Ablation study on surrogate-based ensemble inference strategy.}
\label{fig:ablation}
\end{figure}

The core of our inference strategy lies in utilizing the trained interdiction agent as a surrogate to approximate lower-level evaluations during sampling decoding. To quantify the benefits of the surrogate, we substitute it with a greedy heuristic. The greedy heuristic serves as a simple yet effective estimation for the lower-level response, operating by iteratively removing the facility that causes the maximum loss of coverage. Specifically, we adopt the sampling decoding strategy with $K=128$ as the test configuration. In terms of evaluation metrics, we quantify the solution quality by calculating the objective value gain relative to the greedy decoding strategy. Additionally, we record the inference runtime for each evaluation method, as computational efficiency remains a critical factor for applications. Fig. \ref{fig:ablation} presents the ablation study concerning the inference strategy. In terms of solution quality (left axis), our surrogate-based Inference strategy consistently outperforms the greedy heuristic, yielding higher objective value gains across all tested scales. Regarding efficiency (right axis), while the heuristic's runtime increases drastically with problem scale, our method remains extremely efficient with negligible computational cost. Overall, the ablation study demonstrates that the surrogate-base ensemble inference strategy can enhance the inference performance, while maintaining an efficient inference time.


\section{Conclusion and Future Works}
\label{Conclusion}
\noindent In this article, we propose a dual-agent training framework denoted as DADRL, which can effectively handle the bi-level structure in MCLIP. Specifically, our DADRL consists of a location agent and an interdiction agent, which are trained simultaneously in an adversarial manner following the properties of min-max in the mathematical formulation. To further enhance the inference efficiency, we propose a surrogate-based ensemble inference strategy by utilizing the well-trained interdiction agent as a high-fidelity surrogate during sampling decoding. We evaluate DADRL on both synthetic and real-world datasets, demonstrating that our DADRL outperforms other baselines in both performance and efficiency. To the best of our knowledge, we are the first to apply DRL to solve the BOPs. Therefore, our future work will focus on: 1) extending the framework to other bi-level optimization domains; 2) incorporating complex constraints like capacity limits and stochastic disruptions; and 3) exploring advanced neural architectures to further improve generalization.

\bibliographystyle{IEEEtran}
\bibliography{references.bib}

@article{daskin1983maximum,
  title={A maximum expected covering location model: formulation, properties and heuristic solution},
  author={Daskin, Mark S},
  journal={Transportation Science},
  volume={17},
  number={1},
  pages={48--70},
  year={1983},
  publisher={INFORMS}
}

@article{church1974maximal,
  title={The maximal covering location problem},
  author={Church, Richard and Velle, Charles R},
  journal={Papers in Regional Science},
  volume={32},
  number={1},
  pages={101--118},
  year={1974},
  publisher={Elsevier}
}

@article{schilling1993review,
  title={A review of covering problems in facility location},
  author={Schilling, David A},
  journal={Location Science},
  volume={1},
  pages={25--55},
  year={1993}
}

@incollection{church2018disruption,
  title={Disruption, protection, and resilience},
  author={Church, Richard L and Murray, Alan},
  booktitle={Location Covering Models: History, Applications and Advancements},
  pages={203--227},
  year={2018},
  publisher={Springer}
}

@article{smith2014us,
  title={US risks national blackout from small-scale attack},
  author={Smith, Rebecca},
  journal={Wall Street Journal},
  volume={12},
  year={2014}
}

@article{berman2007facility,
  title={Facility reliability issues in network p-median problems: Strategic centralization and co-location effects},
  author={Berman, Oded and Krass, Dmitry and Menezes, Mozart BC},
  journal={Operations Research},
  volume={55},
  number={2},
  pages={332--350},
  year={2007},
  publisher={INFORMS}
}

@article{snyder2016or,
  title={OR/MS models for supply chain disruptions: A review},
  author={Snyder, Lawrence V and Atan, Z{\"u}mb{\"u}l and Peng, Peng and Rong, Ying and Schmitt, Amanda J and Sinsoysal, Burcu},
  journal={Iie transactions},
  volume={48},
  number={2},
  pages={89--109},
  year={2016},
  publisher={Taylor \& Francis}
}

@article{smith2020survey,
  title={A survey of network interdiction models and algorithms},
  author={Smith, J Cole and Song, Yongjia},
  journal={European Journal of Operational Research},
  volume={283},
  number={3},
  pages={797--811},
  year={2020},
  publisher={Elsevier}
}

@article{drezner1987heuristic,
  title={Heuristic solution methods for two location problems with unreliable facilities},
  author={Drezner, Zvi},
  journal={Journal of the Operational Research Society},
  volume={38},
  number={6},
  pages={509--514},
  year={1987},
  publisher={Taylor \& Francis}
}

@article{lim2010facility,
  title={A facility reliability problem: Formulation, properties, and algorithm},
  author={Lim, Michael and Daskin, Mark S and Bassamboo, Achal and Chopra, Sunil},
  journal={Naval Research Logistics (NRL)},
  volume={57},
  number={1},
  pages={58--70},
  year={2010},
  publisher={Wiley Online Library}
}

@article{li2022general,
  title={A general model and efficient algorithms for reliable facility location problem under uncertain disruptions},
  author={Li, Yongzhen and Li, Xueping and Shu, Jia and Song, Miao and Zhang, Kaike},
  journal={INFORMS Journal on Computing},
  volume={34},
  number={1},
  pages={407--426},
  year={2022},
  publisher={Informs}
}

@article{hogan1986concepts,
  title={Concepts and applications of backup coverage},
  author={Hogan, Kathleen and Revelle, Charles},
  journal={Management Science},
  volume={32},
  number={11},
  pages={1434--1444},
  year={1986},
  publisher={INFORMS}
}

@article{karatas2019analysis,
  title={An analysis of p-median location problem: Effects of backup service level and demand assignment policy},
  author={Karatas, Mumtaz and Yak{\i}c{\i}, Ertan},
  journal={European Journal of Operational Research},
  volume={272},
  number={1},
  pages={207--218},
  year={2019},
  publisher={Elsevier}
}

@article{tao2022location,
  title={Location optimization of urban fire stations considering the backup coverage},
  author={Tao, Liufeng and Cui, Yuqiong and Xu, Yongyang and Chen, Zhanlong and Guo, Han and Huang, Bo and Xie, Zhong},
  journal={International Journal of Environmental Research and Public Health},
  volume={20},
  number={1},
  pages={627},
  year={2022},
  publisher={MDPI}
}

@article{o2011designing,
  title={Designing robust coverage networks to hedge against worst-case facility losses},
  author={O’Hanley, Jesse R and Church, Richard L},
  journal={European Journal of Operational Research},
  volume={209},
  number={1},
  pages={23--36},
  year={2011},
  publisher={Elsevier}
}

@article{church2004identifying,
  title={Identifying critical infrastructure: the median and covering facility interdiction problems},
  author={Church, Richard L and Scaparra, Maria P and Middleton, Richard S},
  journal={Annals of the Association of American Geographers},
  volume={94},
  number={3},
  pages={491--502},
  year={2004},
  publisher={Taylor \& Francis}
}

@article{simaan1973stackelberg,
  title={On the Stackelberg strategy in nonzero-sum games},
  author={Simaan, Marwaan and Cruz Jr, Jose B},
  journal={Journal of Optimization Theory and Applications},
  volume={11},
  number={5},
  pages={533--555},
  year={1973},
  publisher={Springer}
}

@misc{gurobi,
  author = {{Gurobi Optimization, LLC}},
  title = {{Gurobi Optimizer Reference Manual}},
  year = {2024},
  url = {https://www.gurobi.com}
}

@misc{cplex,
  author = {IBM},
  title = {{IBM ILOG CPLEX Optimization Studio}},
  year = {2024},
  url = {https://www.ibm.com/products/ilog-cplex-optimization-studio}
}

@article{camacho2024metaheuristics,
  title={Metaheuristics for bilevel optimization: A comprehensive review},
  author={Camacho-Vallejo, Jos{\'e}-Fernando and Corpus, Carlos and Villegas, Juan G},
  journal={Computers \& Operations Research},
  volume={161},
  pages={106410},
  year={2024},
  publisher={Elsevier}
}

@article{bracken1973mathematical,
  title={Mathematical programs with optimization problems in the constraints},
  author={Bracken, Jerome and McGill, James T},
  journal={Operations Research},
  volume={21},
  number={1},
  pages={37--44},
  year={1973},
  publisher={INFORMS}
}

@article{jeroslow1985polynomial,
  title={The polynomial hierarchy and a simple model for competitive analysis},
  author={Jeroslow, Robert G},
  journal={Mathematical Programming},
  volume={32},
  number={2},
  pages={146--164},
  year={1985},
  publisher={Springer}
}

@article{camacho2014solving,
  title={Solving the bilevel facility location problem under preferences by a Stackelberg-evolutionary algorithm},
  author={Camacho-Vallejo, Jos{\'e}-Fernando and Cordero-Franco, {\'A}lvaro Eduardo and Gonz{\'a}lez-Ram{\'\i}rez, Rosa G},
  journal={Mathematical Problems in Engineering},
  volume={2014},
  number={1},
  pages={430243},
  year={2014},
  publisher={Wiley Online Library}
}

@article{calvete2020AMF,
  title={A matheuristic for solving the bilevel approach of the facility location problem with cardinality constraints and preferences},
  author={Herminia I. Calvete and Carmen Gal{\'e} and Jos{\'e} A. Iranzo and Jos{\'e}-Fernando Camacho-Vallejo and Martha-Selene Casas-Ram{\'i}rez},
  journal={Computers \& Operations Research},
  year={2020},
  volume={124},
  pages={105066}
}

@article{kleinert2020there,
  title={There’s no free lunch: on the hardness of choosing a correct big-M in bilevel optimization},
  author={Kleinert, Thomas and Labb{\'e}, Martine and Plein, Fr{\"{}} ank and Schmidt, Martin},
  journal={Operations Research},
  volume={68},
  number={6},
  pages={1716--1721},
  year={2020},
  publisher={INFORMS}
}

@article{moore1990mixed,
  title={The mixed integer linear bilevel programming problem},
  author={Moore, James T and Bard, Jonathan F},
  journal={Operations Research},
  volume={38},
  number={5},
  pages={911--921},
  year={1990},
  publisher={INFORMS}
}

@article{judice1988solution,
  title={The solution of the linear bilevel programming problem by using the linear complementarity problem},
  author={J{\'u}dice, Joaquim J and Faustino, A},
  journal={Investiga{\c{c}}{\~a}o Operacional},
  volume={8},
  number={1},
  pages={77--95},
  year={1988}
}

@article{bard1990branch,
  title={A branch and bound algorithm for the bilevel programming problem},
  author={Bard, Jonathan F and Moore, James T},
  journal={SIAM Journal on Scientific and Statistical Computing},
  volume={11},
  number={2},
  pages={281--292},
  year={1990},
  publisher={SIAM}
}

@incollection{saharidis2013exact,
  title={Exact solution methodologies for linear and (mixed) integer bilevel programming},
  author={Saharidis, Georgios KD and Conejo, Antonio J and Kozanidis, George},
  booktitle={Metaheuristics for bi-level optimization},
  pages={221--245},
  year={2013},
  publisher={Springer}
}

@article{kleinert2021survey,
  title={A survey on mixed-integer programming techniques in bilevel optimization},
  author={Kleinert, Thomas and Labb{\'e}, Martine and Ljubi{\'c}, Ivana and Schmidt, Martin},
  journal={EURO Journal on Computational Optimization},
  volume={9},
  pages={100007},
  year={2021},
  publisher={Elsevier}
}

@inproceedings{mejia2020surrogate,
  title={A surrogate-assisted metaheuristic for bilevel optimization},
  author={Mej{\'\i}a-de-Dios, Jes{\'u}s-Adolfo and Mezura-Montes, Efr{\'e}n},
  booktitle={Proceedings of the 2020 Genetic and Evolutionary Computation Conference},
  pages={629--635},
  year={2020}
}

@incollection{talbi2013taxonomy,
  title={A taxonomy of metaheuristics for bi-level optimization},
  author={Talbi, El-Ghazali},
  booktitle={Metaheuristics for Bi-level Optimization},
  pages={1--39},
  year={2013},
  publisher={Springer}
}

@incollection{legillon2013cobra,
  title={Cobra: A coevolutionary metaheuristic for bi-level optimization},
  author={Legillon, Fran{\c{c}}ois and Liefooghe, Arnaud and Talbi, El-Ghazali},
  booktitle={Metaheuristics for Bi-level Optimization},
  pages={95--114},
  year={2013},
  publisher={Springer}
}

@article{panin2014bilevel,
  title={Bilevel competitive facility location and pricing problems},
  author={Panin, Artem A and Pashchenko, Mikhail G and Plyasunov, Aleksandr V},
  journal={Automation and Remote Control},
  volume={75},
  number={4},
  pages={715--727},
  year={2014},
  publisher={Springer}
}

@article{esfahani2022optimal,
  title={Optimal pricing for bidirectional wireless charging lanes in coupled transportation and power networks},
  author={Esfahani, Hossein Nasr and Liu, Zhaocai and Song, Ziqi},
  journal={Transportation Research Part C: Emerging Technologies},
  volume={135},
  pages={103419},
  year={2022},
  publisher={Elsevier}
}

@article{li2022hybrid,
  title={A hybrid heuristic approach with adaptive scalarization for linear semivectorial bilevel programming and its application},
  author={Li, Hong and Zhang, Li},
  journal={Memetic Computing},
  volume={14},
  number={4},
  pages={433--449},
  year={2022},
  publisher={Springer}
}

@article{hayashi2023bilevel,
  title={Bilevel optimization model for sizing of battery energy storage systems in a microgrid considering their economical operation},
  author={Hayashi, Ryosuke and Takano, Hirotaka and Nyabuto, Welma Mogiti and Asano, Hiroshi and Nguyen-Duc, Tuyen},
  journal={Energy Reports},
  volume={9},
  pages={728--737},
  year={2023},
  publisher={Elsevier}
}

@article{ziar2023efficient,
  title={An efficient environmentally friendly transportation network design via dry ports: a bi-level programming approach},
  author={Ziar, Elham and Seifbarghy, Mehdi and Bashiri, Mahdi and Tjahjono, Benny},
  journal={Annals of Operations Research},
  volume={322},
  number={2},
  pages={1143--1166},
  year={2023},
  publisher={Springer}
}

@article{peng2022research,
  title={Research on location-routing problem of maritime emergency materials distribution based on bi-level programming},
  author={Peng, Zhongxiu and Wang, Cong and Xu, Wenqing and Zhang, Jinsong},
  journal={Mathematics},
  volume={10},
  number={8},
  pages={1243},
  year={2022},
  publisher={MDPI}
}

@article{lu2022bilevel,
  title={A bilevel whale optimization algorithm for risk management scheduling of information technology projects considering outsourcing},
  author={Lu, Fuqiang and Yan, Tongren and Bi, Hualing and Feng, Ming and Wang, Suxin and Huang, Min},
  journal={Knowledge-Based Systems},
  volume={235},
  pages={107600},
  year={2022},
  publisher={Elsevier}
}

@article{zhou2023bilevel,
  title={Bilevel memetic search approach to the soft-clustered vehicle routing problem},
  author={Zhou, Yangming and Kou, Yawen and Zhou, MengChu},
  journal={Transportation Science},
  volume={57},
  number={3},
  pages={701--716},
  year={2023},
  publisher={INFORMS}
}

@article{said2022discretization,
  title={Discretization-based feature selection as a bilevel optimization problem},
  author={Said, Rihab and Elarbi, Maha and Bechikh, Slim and Coello, Carlos Artemio Coello and Said, Lamjed Ben},
  journal={IEEE Transactions on Evolutionary Computation},
  volume={27},
  number={4},
  pages={893--907},
  year={2022},
  publisher={IEEE}
}

@article{chen2022integrated,
  title={Integrated optimization of transfer station selection and train timetables for road--rail intermodal transport network},
  author={Chen, Xinghan and Zuo, Tianshuai and Lang, Maoxiang and Li, Shiqi and Li, Siyu},
  journal={Computers \& Industrial Engineering},
  volume={165},
  pages={107929},
  year={2022},
  publisher={Elsevier}
}

@inproceedings{kool2018attention,
  author       = {Wouter Kool and
                  Herke van Hoof and
                  Max Welling},
  title        = {Attention, Learn to Solve Routing Problems!},
  booktitle    = {7th International Conference on Learning Representations, {ICLR} 2019,
                  New Orleans, LA, USA, May 6-9, 2019},
  year         = {2019},
}

@inproceedings{miao2024deep,
  title={Deep Reinforcement Learning for Multi-Period Facility Location: pk-median Dynamic Location Problem},
  author={Miao, Changhao and Zhang, Yuntian and Wu, Tongyu and Deng, Fang and Chen, Chen},
  booktitle={Proceedings of the 32nd ACM International Conference on Advances in Geographic Information Systems},
  pages={1--11},
  year={2024}
}

@article{liang2024sponet,
  title={Sponet: solve spatial optimization problem using deep reinforcement learning for urban spatial decision analysis},
  author={Liang, Haojian and Wang, Shaohua and Li, Huilai and Zhou, Liang and Chen, Hechang and Zhang, Xueyan and Chen, Xu},
  journal={International Journal of Digital Earth},
  volume={17},
  number={1},
  pages={2299211},
  year={2024},
  publisher={Taylor \& Francis}
}

@article{zhong2024recovnet,
  title={ReCovNet: Reinforcement learning with covering information for solving maximal coverage billboards location problem},
  author={Zhong, Yang and Wang, Shaohua and Liang, Haojian and Wang, Zhenbo and Zhang, Xueyan and Chen, Xi and Su, Cheng},
  journal={International Journal of Applied Earth Observation and Geoinformation},
  volume={128},
  pages={103710},
  year={2024},
  publisher={Elsevier}
}

@article{wang2023deepmclp,
  title={DeepMCLP: Solving the MCLP with Deep Reinforcement Learning for Urban Spatial},
  author={Wang, Shaohua and Liang, Haojian and Zhong, Yang},
  journal={Soft Computing},
  year={2023}
}

@article{miao2025end,
  title={An End-to-End Learning Approach for Solving Capacitated Location-Routing Problems},
  author={Miao, Changhao and Zhang, Yuntian and Wu, Tongyu and Deng, Fang and Chen, Chen},
  journal={arXiv preprint arXiv:2511.02525},
  year={2025}
}

@article{huang2025deep,
  title={A deep reinforcement learning method for solving Two-Echelon Location-Routing Problem},
  author={Huang, Shuo and Wu, Yaoxin and Cao, Zhiguang and Zhang, Xuexi},
  journal={Computers \& Operations Research},
  pages={107210},
  year={2025},
  publisher={Elsevier}
}

@article{chen2023attention,
  title={An attention model with multiple decoders for solving p-Center problems},
  author={Chen, Xu and Wang, Shaohua and Li, Huilai and Liang, Haojian and Li, Ziqiong and Lu, Hao},
  journal={International Journal of Applied Earth Observation and Geoinformation},
  volume={125},
  pages={103526},
  year={2023},
  publisher={Elsevier}
}

@inproceedings{goodfellow2014generative,
  title={Generative adversarial nets},
  author={Goodfellow, Ian and Pouget-Abadie, Jean and Mirza, Mehdi and Xu, Bing and Warde-Farley, David and Ozair, Sherjil and Courville, Aaron and Bengio, Yoshua},
  booktitle={Advances in Neural Information Processing Systems},
  volume={27},
  pages={2672--2680},
  year={2014}
}

@article{pereira2007column,
  title={A column generation approach for the maximal covering location problem},
  author={Pereira, Marcos Antonio and Lorena, Luiz Antonio Nogueira and Senne, Edson Luiz Fran{\c{c}}a},
  journal={International Transactions in Operational Research},
  volume={14},
  number={4},
  pages={349--364},
  year={2007},
  publisher={Wiley Online Library}
}

@article{GM,
  title={A heuristic program for locating warehouses},
  author={Kuehn, Alfred A and Hamburger, Michael J},
  journal={Management Science},
  volume={9},
  number={4},
  pages={643--666},
  year={1963},
  publisher={INFORMS}
}

@article{GE,
  title={A fast algorithm for the greedy interchange for large-scale clustering and median location problems},
  author={Whitaker, RA0527},
  journal={INFOR: Information Systems and Operational Research},
  volume={21},
  number={2},
  pages={95--108},
  year={1983},
  publisher={Taylor \& Francis}
}

@article{GI,
  title={A more efficient heuristic for solving large p-median problems},
  author={Densham, Paul J and Rushton, Gerard},
  journal={Papers in Regional Science},
  volume={71},
  number={3},
  pages={307--329},
  year={1992},
  publisher={Springer}
}

@article{Stingy,
  title={Warehouse location under continuous economies of scale},
  author={Feldman, EFATL and Lehrer, FA and Ray, TL},
  journal={Management Science},
  volume={12},
  number={9},
  pages={670--684},
  year={1966},
  publisher={INFORMS}
}

@article{AS,
  title={On the location of supply points to minimize transport costs},
  author={Maranzana, FE},
  journal={Journal of the Operational Research Society},
  volume={15},
  number={3},
  pages={261--270},
  year={1964},
  publisher={Taylor \& Francis}
}

@article{GA,
  title={The large scale maximal covering location problem},
  author={Zarandi, MH Fazel and Davari, Soheil and Sisakht, SA Haddad},
  journal={Scientia Iranica},
  volume={18},
  number={6},
  pages={1564--1570},
  year={2011},
  publisher={Elsevier}
}

@article{SA,
  title={Maximal covering location problem (MCLP) with fuzzy travel times},
  author={Davari, Soheil and Zarandi, Mohammad Hossein Fazel and Hemmati, Ahmad},
  journal={Expert Systems with Applications},
  volume={38},
  number={12},
  pages={14535--14541},
  year={2011},
  publisher={Elsevier}
}

@article{TS,
  title={The minimum weighted covering location problem with distance constraints},
  author={Berman, Oded and Huang, Rongbing},
  journal={Computers \& Operations Research},
  volume={35},
  number={2},
  pages={356--372},
  year={2008},
  publisher={Elsevier}
}

@article{VNS,
  title={A variable neighborhood search for the budget-constrained maximal covering location problem with customer preference ordering},
  author={Mrkela, Lazar and Stanimirovi{\'c}, Zorica},
  journal={Operational Research},
  volume={22},
  number={5},
  pages={5913--5951},
  year={2022},
  publisher={Springer}
}

\end{document}